\documentclass[journal]{IEEEtran}

\usepackage{graphics,epsfig} 
\usepackage{amsmath,amsfonts,amssymb,bm}
\usepackage{subfigure,cases,multirow}
\usepackage{algorithm,algpseudocode,algorithmicx}

\usepackage[colorlinks,citecolor=blue,urlcolor=black,linkcolor=blue]{hyperref}
\usepackage{accents}
\usepackage[displaymath,mathlines,switch]{lineno}
\usepackage[mathscr]{eucal}
\usepackage{booktabs}

\makeatletter
\def\wideubar{\underaccent{{\cc@style\underline{\mskip10mu}}}}
\def\ubar{\underaccent{{\cc@style\underline{\mskip6mu}}}}
\makeatother

\makeatletter
\def\widebar{\accentset{{\cc@style\underline{\mskip10mu}}}}
\makeatother

\ifCLASSOPTIONcompsoc
  \usepackage[nocompress]{cite}
\else
  \usepackage{cite}
\fi

\ifCLASSINFOpdf
\else
\fi
\def\bR{\mathbb R}
\def\bS{\mathbb S}

\def\bZ{\mathbb Z}
\def\bM{\mathbb M}

\def\rQ{\mathscr Q}

\def\cE{\mathcal E}
\def\cD{\mathcal D}

\def\cV{\mathcal V}
\def\cL{\mathcal L}
\def\cU{\mathcal U}
\def\cC{\mathcal C}
\def\cM{\mathcal M}
\def\cG{\mathcal G}
\def\cF{\mathcal F}
\def\cE{\mathcal E}
\def\cJ{\mathcal J}
\def\cW{\mathcal W}
\def\cP{\mathcal P}
\def\cN{\mathcal N}
\def\cS{\mathcal S}
\def\cH{\mathcal H}

\def\fy{\mathbf y}
\def\fz{\mathbf z}
\def\fx{\mathbf x}
\def\fq{\mathbf q}
\def\fa{\mathbf a}
\def\fb{\mathbf b}
\def\fs{\mathbf s}
\def\fu{\mathbf u}
\def\fv{\mathbf v}
\def\fn{\mathbf n}
\def\fe{\mathbf e}

\DeclareMathOperator\Lip{Lip}
\DeclareMathOperator\sign{Sign}
\DeclareMathOperator\Id{I_d}

\DeclareMathOperator\Trial{Trial}
\DeclareMathOperator\Accepted{Accepted}
\DeclareMathOperator\StoppingFlag{\textsc{StoppingFlag}}
\DeclareMathOperator\Satisfactory{\textsc{AdmissibleNeigh}}

\newlength\savewidth
\newcommand\shline{\noalign{\global\savewidth\arrayrulewidth\global\arrayrulewidth 1.0pt}\hline\noalign{\global\arrayrulewidth\savewidth}}

\newlength\savedwidth

\begin{document}
\title{Geodesic Paths for Image Segmentation with Implicit Region-based Homogeneity Enhancement}
\author{Da~Chen, Jian Zhu, Xinxin Zhang, Minglei Shu and Laurent D. Cohen, ~\IEEEmembership{Fellow,~IEEE}
\thanks{Da Chen and Minglei Shu are with Shandong Artificial Intelligence Institute, Qilu University of Technology (Shandong Academy of Sciences), China.~(e-mails:~dachen.cn@hotmail.com;\,shuml@sdas.org) (Minglei Shu is the corresponding author)}
\thanks{Jian Zhu is with Department of Radiation Oncology Physics \& Technology, Shandong Cancer Hospital affiliated to Shandong First Medical University, Jinan, China.}
\thanks{Xinxin Zhang is with the School of Software, Shandong University,  Jinan, China.}
\thanks{Laurent D. Cohen is with University Paris Dauphine, PSL Research University, CNRS, UMR 7534, CEREMADE, 75016 Paris, France.}
}
\markboth{Journal of \LaTeX\ Class Files,~Vol.~16, No.~8, August~2020}
{Shell \MakeLowercase{\textit{et al.}}: Bare Demo of IEEEtran.cls for Journals}
\maketitle

\begin{abstract}
Minimal paths are regarded as a powerful and efficient tool for boundary detection and image segmentation due to its global optimality and the well-established numerical solutions such as fast marching method. In this paper, we introduce a flexible interactive image segmentation model based on the Eikonal partial differential equation (PDE) framework in conjunction with region-based homogeneity enhancement. A key ingredient in the introduced model is the construction of local  geodesic metrics, which are capable of integrating anisotropic and asymmetric edge features, implicit region-based homogeneity features and/or curvature regularization. The incorporation of the region-based  homogeneity features into the metrics considered relies on an implicit representation of these features, which is one of the contributions of this work. Moreover, we also introduce a way to build simple closed contours as the concatenation of two disjoint open curves. Experimental results prove that the proposed model indeed outperforms state-of-the-art minimal paths-based image segmentation approaches. 
\end{abstract}


\IEEEpeerreviewmaketitle

\section{Introduction}
\label{sec:introduction}
Image segmentation is a fundamental task in a great variety of applications arising in the fields of computer vision and medical imaging.  The segmentation approaches based on the energy minimization theorems, such as the variational methods or the graph-based methods,  have demonstrated their strong capacity of coping with various challenging image segmentation issues. Among them, the interactive segmentation algorithms in conjunction with user intervention and priors are able to provide a reliable and efficient way for separating foreground regions of interest from image domain. 

The interventions from user often provide necessary information to initialize the interactive segmentation approaches, or impose effective constraints to encourage reasonable and accurate segmentations. In many segmentation approaches, user interactions can be constructed by loosely drawing scribbles associated to different regions. These scribbles serve as initial seeds for image segmentation. Models relying on a graph-based optimization scheme frequently utilize such an interactive fashion as introduced in~\cite{boykov2006grapy,grady2006random,couprie2011power,li2004lzay}, for which an image is modeled as a graph collecting a set of nodes and edges.   The Voronoi diagram-based segmentation approaches~\cite{arbelaez2004energy,bai2009geodesic,chen2018fast} implement the image domain partitioning  through Voronoi regions and the corresponding Voronoi index maps, where the user-provided scribbles serve as the sets of source points for the computation of minimal weighted distances and for the propagation of region labels. In~\cite{gao2012interactive}, these scribbles were treated as subregions of the image domain, from which statistical models fitting to the image intensity distributions in the target regions are created. This is also the case for the selective segmentation models~\cite{spencer2019parameter,nguyen2012robust}, which exploited user-provided scribbles to extract statistical priors of image features. 

Active contour approaches~\cite{kass1988snakes} have proven their ability in addressing a wide variety of image segmentation problems. In their basic formulation, the segmentation procedure can be carried out by deforming initial curves driven by  suitable gradient flows. These initial curves can be placed close to the targets, thus able to identify specific target regions from complicated backgrounds, especially for these models relying on local image features such as image gradients~\cite{kass1988snakes,xu1998snakes,cohen1991active,cohen1993finite,caselles1997geodesic,malladi1995shape} and local region-based homogeneity penalization~\cite{li2008minimization,brox2009local}. Thanks to the energy minimization framework, the geometric priors such as Euclidean curve length and the elastica energy can be naturally taken into account for finding favorable segmentations.
The  interventions created by clicking several points along the boundary of interest often serve as the user input for paths-based interactive image segmentation models. In general, these models usually exploit  closed contours to delineate target boundaries, each of which can be sought via a set of relevant piecewise minimal cost paths. Given suitable cost functions for curve arcs, these minimal cost paths can be efficiently tracked  either in a discrete setting~\cite{miranda2012riverbed} or in a continuous PDE framework~\cite{cohen1997global}.  

Shape priors can be naturally incorporated into segmentation models in a   energy minimization framework such as active contours~\cite{cremers2002diffusion,bresson2006variational,leventon2000statistical, chan2005level,klodt2011convex}, allowing to encourage segmented object regions to satisfy the constraints induced from the given shape priors. In addition, recent segmentation approaches impose that the segmented regions are convex~\cite{yan2020convexity,luo2019convex,gorelick2016convexity,royer2016convexity} or star convexity~\cite{veksler2008star,vicente2008graph,yuan2012efficient}, which are capable of generating promising image segmentation results in many challenging scenarios. The list of the literature reviewed above is obviously not exhaustive and other interesting and efficient image segmentation approaches may include the learning-based models such as~\cite{ma2020learning,wang2018deepigeos,zhang2020exploring,zhang2021automatic,zhang2021cross}.
 In the following, we concentrate on the minimal geodesic path approaches under the framework of Eikonal PDEs. 
 
 \begin{figure}[t]
 \centering
\includegraphics[width=8.5cm]{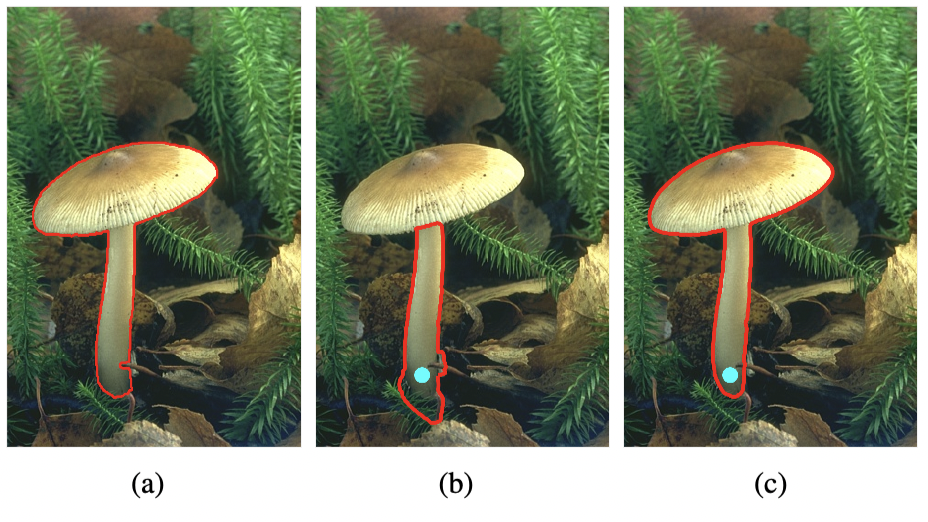}
 \caption{An example for the illustration of the advantages of the proposed geodesic paths-based segmentation model. (\textbf{a}) An original image, where the red line indicates the ground truth. (\textbf{b}) and (\textbf{c}) Segmentation contours from the edge-based circular geodesic model~\cite{appleton2005globally} and the proposed model, respectively}
 \label{fig_Example}
 \end{figure}

\subsection{Geodesic Paths-based Image Segmentation Models} 
The original snakes model~\cite{kass1988snakes} invoked a non-intrinsic functional that depends on the parameterization of the evolving curves. The geodesic active contour models~\cite{caselles1997geodesic,yezzi1997geometric,kimmel2003regularized} remove the dependency on curve parameterization. These geometric approaches made use of weighted curve lengths as energy functionals, which are  measured via  a type of Riemannian metrics. Contrary to  the snakes model~\cite{kass1988snakes}  using parameterized curves, the curve evolution in these geometric approaches can be implemented in a level set formulation~\cite{osher1988fronts}. However, as an important shortcoming, it is difficult for these geometric active contour models to find the global minimum of the corresponding weighted curve length. As a consequence, the image segmentations are  sensitive to the initialization. In order to overcome this issue, Cohen and Kimmel~\cite{cohen1997global} introduced a minimal geodesic model based on the Eikonal PDE framework, where a globally minimizing curve is a geodesic path associated to a Riemannian metric.  

The minimal geodesic models~\cite{peyre2010geodesic} are quite efficient for image segmentation applications, due to the well-studied numerical schemes such as the Fast marching method, and the global optimality. Along this research line, most of the relevant approaches~\cite{benmansour2009fast,mille2009geodesically,mille2015combination} attempted to construct simple and closed contours leveraging  geodesic paths. Cohen and Kimmel introduced a saddle points detection approach for image segmentation~\cite{cohen1997global}. The initialization is a single point located in the boundary of interest, from which a closed contour can be generated to describe the target boundary. This saddle point detection method was then adopted by~\cite{appia2011active,mille2015combination} for interactive image segmentation in conjunction with a set of prescribed points at the target boundary. However, the geodesic paths in these models rely only on the edge-based features by essence, despite the use of region-based homogeneity terms for finding the final segmentations. This issue is addressed in~\cite{chen2016finsler,chen2019active}, where a Randers minimal path model was exploited as a solution to the region-based active contour problems. Unfortunately, this model does not take into account curvature regularization when computing minimal paths. 

Instead of placing source points at the target boundary, the circular geodesic model~\cite{appleton2005globally} exploits a fixed point inside the target region as initialization to set up the segmentation  algorithm. However, neither the region-based terms nor the curvature regularization were  exploited for segmentation. 
A simple closed geodesic path is  extracted  in conjunction of a particular cut placed in the image domain.  In this paper, we propose a new  geodesic paths-based image segmentation model relying on a dual-cut scheme. The proposed model differs to the original circular geodesic model~\cite{appleton2005globally} mainly at the construction of the local geodesic metrics and at the generation of simple closed contours.  Specifically, the geodesic metrics considered are able to encode the region-based homogeneity features and priors for image segmentation. As mentioned above, the original circular geodesic model~\cite{appleton2005globally} still falls into the \emph{edge-only} limitation, such that the resulting  segmentation curves may fail to capture the whole target region, as illustrated in Fig.~\ref{fig_Example}b. In contrast,  one can see that the integration of those beneficial features in the proposed model can overcome such an issue, as depicted in Fig.~\ref{fig_Example}c. In this experiment, Fig.~\ref{fig_Example}a illustrates the original image where the red line indicates the ground truth contour.

\subsection{Contributions and Paper Structure}
 The main contributions of this paper are three folder. 
 \begin{itemize}
 	\item \emph{Geodesic metrics implicitly encoding region-based homogeneity features}. We introduce a new geodesic metric construction method for minimal path computation with application to interactive image segmentation. The  geodesic metrics considered can be decomposed into a scalar-valued function which encodes region-based homogeneity features, and a type of Finsler metrics which involve either the asymmetric image gradient features or the curvature regularization.
 	\item \emph{Dual-cut Scheme for closed contour construction}. Inspired by the circular geodesic model~\cite{appleton2005globally}, we propose a new geodesic paths-based interactive image segmentation model relying on a point that is located inside the target region. The final segmentation contours are generated by the concatenation of two disjoint paths derived from a new dual-cut scheme. 
 	\item \emph{Flexible implementation of user intervention}. We show that the scribbles provided by the user can be easily incorporated into the proposed segmentation method. These scribbles are regarded as barriers to prevent the geodesic paths from crossing over unexpected positions, allowing a flexible implementation of the proposed model for efficient interactive image segmentation.
 \end{itemize}
 
The remaining of this paper is organized as follows. In Section~\ref{sec_BG}, we introduce the background on a generic minimal geodesic model and on the computation of image features.  The main contributions are presented in Sections~\ref{sec_principle} and~\ref{sec_DualCut}. Specifically, Section~\ref{sec_principle} introduces a new metric construction method which integrates with image edge-based features, implicit region-based homogeneity features  and/or the curvature regularization term. Moreover, Section~\ref{sec_DualCut} presents a dual-cut scheme for the extraction of simple closed curves.  Experimental results and conclusion are presented in Sections~\ref{sec_Exp} and~\ref{sec_Conclusion}, respectively.

\section{BackGround}
\label{sec_BG}
\subsection{Minimal Paths}
\label{subsec_MP}
Tracking continuous curves to depict interesting image features is a fundamental problem posed in the field of image analysis.  Cohen and Kimmel introduced\cite{cohen1997global} an elegant minimal path solution to that problem based on the Eikonal PDE framework, yielding a broad variety of successful applications.

Let $\bM\subset\bR^n$ be an open and bounded domain of dimension $n=2,3$. Basically, a core ingredient for minimal path models is the energy for a curve $\gamma$, defined as the weighed curve length of $\gamma$ associated to a geodesic metric $\cF:\bM\times\bR^n\to\bR_0^+$. At each fixed point $\fx\in\bM$, the metric can be denoted by $\cF(\fx,\fu)=F_{\fx}(\fu)$, where $F_{\fx}(\fu)$ is a 1-homogeneous and convex function. 
In the original minimal path model~\cite{cohen1997global}, the weighted length of a curve is measured by isotropic Riemannian metrics, which are independent to the curve tangents $\gamma^\prime$ (i.e. the first-order derivative of $\gamma$).  In general, a Finsler metric $\cF(\fx,\fu)$ is allowed to be asymmetric and anisotropic with respect to its second argument at some point $\fx\in\bM$~\cite{melonakos2008finsler}. Typical examples of Finsler metric may involve the Randers metrics~\cite{randers1941asymmetrical,chen2016finsler,chen2017global} and the asymmetric quadratic metrics~\cite{chen2018asymmetric,duits2018optimal,mirebeau2018fast}.  

The weighted curve length of a Lipschitz continuous curve $\gamma:[0,1]\to\bM$, measured using a general Finsler metric $\cF$, can be formulated by
\begin{equation}
\label{eq:CL}
\cL_{\cF}(\gamma):=\int_0^1\cF(\gamma(u),\gamma^\prime(u))du.	
\end{equation}
Given a fixed source point $\fs\in\bM$, globally minimizing the weighted curve length~\eqref{eq:CL} between $\fs$ and an arbitrary target point $\fx\in\bM$ yields a geodesic distance map $\cU_\fs:\bM\to\bR_0^+$
\begin{equation}
\label{eq_MinimalLength}
\cU_{\fs}(\fx)=\inf_{\gamma\in\Lip([0,1],\bM)}~\big\{\cL_\cF(\gamma);\gamma(0)=\fs,\gamma(1)=\fx\big\},
\end{equation}
where $\Lip([0,1],\bM)$ is the set of all Lipschitz curves $\gamma:[0,1]\to\bM$. A geodesic path linking from the source point $\fs$ to a target point $\fx$ is a globally minimizing curve $\cG_{\fs,\fx}\in\Lip([0,1],\bM)$ such that its weighted curve length is equivalent to the geodesic distance $\cU_\fs(\fx)$, i.e.
\begin{equation}
\label{eq_GeoCurve}
\cG_{\fs,\fx}=\underset{\gamma\in\Lip([0,1],\bM)}{\arg\min}\big\{\cL_\cF(\gamma);\gamma(0)=\fs,\gamma(1)=\fx\big\}.
\end{equation}
The geodesic distance map $\cU_\fs$ associated to a Finsler metric $\cF$ admits the unique viscosity solution to a generalized Eikonal PDE, or a static Hamilton-Jacobi PDE~\cite{mirebeau2014efficient}, which reads
\begin{equation}
\label{eq_FinslerEikonal}
\begin{cases}
\displaystyle\sup_{\fv\neq\mathbf{0}}\frac{\langle\nabla \cU_\fs(\fx),\fv\rangle}{\cF(\fx,\fv)}=1,~\forall \fx\in\Omega\backslash\{\fs\},\\
\cU_\fs(\fs)=0,
\end{cases}
\end{equation}
where $\langle\fu_1,\fu_2\rangle=\fu_1^T\fu_2$ denotes the standard Euclidean scalar product of two vectors $\fu_1,\,\fu_2\in\bR^n$.

Tracing a geodesic path $\cG_{\fs,\fx}$, as defined in Eq.~\eqref{eq_GeoCurve}, can be implemented by re-parameterizing the solution $\cG$ to a gradient descent ordinary differential equation (ODE) such that  $\cG(0)=\fx$, and for $u>0$
\begin{equation}
\label{eq_ODE}
\cG^\prime(u)=-\,\underset{\|\fv\|=1}{\arg\max}\frac{\langle\nabla \cU_\fs(\cG(u)),\fv\rangle}{\cF(\cG(u),\fv)}.
\end{equation}
The back-tracking procedure~\eqref{eq_ODE} will be terminated once the source point $\fs$ is reached. Numerically, the gradient descent ODE can be  solved by the efficient scheme proposed in~\cite{mirebeau2019riemannian}.

In the remaining of this paper, we explore two types of Finsler geodesic metrics to deal with the 2D image segmentation problem, as introduced in Sections~\ref{subsec_SpatialMetric} and~\ref{subsec_Curvature}. The first type of considered metrics is constructed using image features only. In this case, one has $\bM:=\Omega$, where $\Omega\subset\bR^2$ stands for an open and bounded \emph{image domain}. The second type of metrics invokes a tool of  orientation lifting to to track curvature-regularized geodesic paths. Accordingly, these geodesics paths are established over an orientation-lifted space $\bM:=\Omega\times \bS^1$, where $\bS^1:=\bR\backslash(2\pi\bZ)$ denotes the orientation space with a periodic boundary condition.

\subsection{Region-based Active Contour Models}
\label{subsec_RAC}
The region-based active contour models\cite{chan2001active,cremers2007review,zhu1996region,dubrovina2015multi} implement the image segmentation by minimizing an energy functional $\Psi$ with respect to  \emph{closed} curves $\cC:[0,1]\to\Omega$ 
\begin{equation}
\label{eq_ACEnergy}
\Psi(\cC)=\cE(\cC)+\eta\cL(\cC),
\end{equation}
where $\eta\in\bR^+$ is a parameter that controls the relative importance between the region-based term $\cE$ and the regularization term $\cL$. Specifically, the term $\cL(\cC)$ can be set as either the Euclidean curve length  of $\cC$ or as the weighted curve length~\eqref{eq:CL}. The region-based homogeneity penalization is encoded in the term $\cE(\cC)$, where typical examples may include the region competition models~\cite{zhu1996region,chan2001active}, the pairwise similarity models~\cite{jung2012nonlocal,sumengen2006graph} and the Bhattacharyya coefficient-based model~\cite{michailovich2007image}.

 In the context of two-phase segmentation, the curve $\cC$ partitions the image domain $\Omega$ into two regions $R$ and $\Omega\backslash R$, where we assume $R$ is enclosed by $\cC$. In the following, $R$ is also referred to as a shape. We take the piecewise constants model~\cite{chan2001active,chan2000active} as an instance, for which the corresponding regional term $\cE$ reads as
\begin{equation}
\label{eq_PConst}
\cE(\cC)=\int_{R}\|I(\fx)-\mathbf{c}_1\|^2d\fx+\int_{\Omega\backslash R}\|I(\fx)-\mathbf{c}_2\|^2d\fx,	
\end{equation}
where $I:\Omega\to\bR^m$ is an image, with $m=1$ and  $m=3$ corresponding to gray level images and color images, respectively.
The scalar value $\mathbf{c}_1=(c_{1,1},\cdots,c_{m,1})$ (resp. $\mathbf{c}_2=(c_{1,2},\cdots,c_{m,2})$) stands for the mean intensities of the corresponding channel of $I$ within the subregion $R$ (resp. the subregion $\Omega\backslash R$), i.e., 
\begin{equation}
\label{eq_Means}
c_{k,1}:=\frac{\int_{R}I_k(\fx)d\fx}{\int_{R}d\fx}, \quad c_{k,2}:=\frac{\int_{\Omega\backslash R}I_k(\fx)d\fx}{\int_{\Omega\backslash R}d\fx},
\end{equation}
for $k=1,\cdots,m$.
\subsection{Edge-based Features from Image Gradients}
\label{subsec_EdgeFeatures}
The extraction of image edge appearance and anisotropy features very often relies on the image gradients. We adopt the model introduced in~\cite{sochen1998vision} for the computation of the gradients of a color image $I=(I_1,I_2,I_3)$. This is implemented by invoking the Jacobian matrices of the smoothed image  $G_\sigma\ast I$, where $G_\sigma$ is a Gaussian kernel of standard derivation $\sigma$ and where `$\ast$' stands for the convolution operator. 
As in~\cite{sochen1998vision}, we first build a Jacobian matrix $\cJ(\fx)=(\nabla G_\sigma \ast I)(\fx)$ of size $2\times 3$, which is comprised of the smoothed $x$-derivatives $\partial_x G_\sigma \ast I_i$ and $y$-derivatives $\partial_y G_\sigma\ast I_i$ for $i=1,2,3$. 

\subsubsection{Edge Appearance Features}
Based on the Jacobian matrices $\cJ$, the image edge appearance features can be carried out by the Frobenius norms of $\cJ$, which reads 
\begin{equation}
\label{eq_FrobNorms}
\tilde{g}(\fx)=\left(\sum_i^3\,\|(\nabla G_\sigma\ast I_i)(\fx)\|^2\right)^{1/2}.	
\end{equation}
In practice, we normalize the values of the scalar-valued function $\tilde{g}$ to the range $[0,1]$ by defining
\begin{equation}
\label{eq_Magnitude}	
g(\fx)=\frac{\tilde{g}(\fx)}{\displaystyle\sup_{\fy\in\Omega}\|\tilde{g}(\fy)\|},\quad \forall \fx\in\Omega.
\end{equation}
\subsubsection{Edge Anisotropic Features}
Let $\bS_2^+$ stand for the set collecting all positive definite symmetric tensors of size $2\times2$. In order to compute the edge anisotropy features, we take into account a tensor field $\cW\in\bS_2^+$ which can be expressed as~\cite{sochen1998vision} 
\begin{equation}
\label{eq_W}
\cW(\fx)=\cJ(\fx)\cJ(\fx)^T+\Id,
\end{equation}
where $\Id$ is the identity of size $2\times2$. The edge anisotropic features are encoded in the matrices $\cW(\fx)$.

Notice that for a gray level image $I:\Omega\to\bR$,  one has $\cJ(\fx)=(\partial_x G_\sigma\ast I,\partial_y G_\sigma\ast I)^T$, and the corresponding tensor $\cW(\fx)$ can be  still computed using Eq.~\eqref{eq_W}.

\section{Geodesic Paths with Implicit Region-based Homogeneity Enhancement}
\label{sec_principle}
The core contributions of this paper lie at the introduction of a new minimal geodesic model for efficient interactive image segmentation. Basically, the proposed model is mainly comprised of two ingredients: (i) the computation of local geodesic metrics encoding image features and/or curvature regularization, and (ii) the construction of simple closed curves made up of geodesic paths. Both of them require a point $\fz\in\Omega$ to provide reliable user intervention such that the point  $\fz$, referred to as a  landmark point, is supposed to be placed inside the target region. 

In this section, we focus on the computation of local geodesic metrics which implicitly encode the region-based homogeneity features. In the context of image segmentation, image gradients involving both edge appearance and anisotropy features are very often implemented to define object boundaries. However,  exploiting only image gradients for building metrics is usually insufficient to find favorable segmentation results in many  complex scenarios, as illustrated in Fig.~\ref{fig_Example}.  In order to address this issue, we consider a geodesic metric $\rQ_z$ associated to the landmark point $\fz$,  which can be expressed as
\begin{equation}
\label{eq_targetMetric}
\rQ_\fz(\fx,\fu):=\psi_\fz(\fx)\cF(\fx,\fu),	
\end{equation}
where $\psi_\fz:\Omega\to\bR^+$ is a scalar-valued weighted function that encodes the region-based homogeneity information.  The second term $\cF$ in the right side of Eq.~\eqref{eq_targetMetric} is a Finsler metric. Moreover, the metric $\cF$ allows to incorporate the curvature-dependent length terms as regularization. The construction for $\cF$ will be detailed in Sections~\ref{subsec_SpatialMetric} and~\ref{subsec_Curvature}. 

\subsection{Extracting Implicit Region-based Homogeneity Features}
\label{subsec_ImplicitFeatures}

In a great variety of region-based active contour models, the motion of  planar closed curves $\cC:[0,1]\to\Omega$ satisfies the following evolution equation
\begin{equation}
\label{eq_GradientFlow}
\frac{\partial\cC(u)}{\partial t}=\xi(\cC(u))\cN(u),	
\end{equation}
where $\cN$ is the \emph{inward} unit normal to $\cC$ and  $\xi:\Omega\to\bR$ is a velocity function. As discussed in the literature~\cite{zhu1996region,tsai2001curve}, the flow~\eqref{eq_GradientFlow} can be exploited to minimize a region-based energy functional of a form 
\begin{equation}
\label{eq_GeneralFunctional}
E(\cC)=\int_R\xi(\fx)d\fx=\int_\Omega\xi(\fx)\chi_R(\fx)d\fx,
\end{equation}
where $R\subset\Omega$ is the interior region of $\cC$ and $\chi_R:\Omega\to\{0,1\}$ stands for the characteristic function of $R$. The velocity $\xi$ can be chosen as the $L^1$ shape gradient of a region-based functional at $\chi_{R_0}$, where $R_0\subset\Omega$ is referred to as the initial shape. We refer to~\cite{chen2019eikonal} for more details on the $L^1$ shape gradient.  Notice that $R_0$ is supposed to contain the  point $\fz$, which can be built by running a front propagation expanding from $\fz$, as described in Section~\ref{subsec_Initializations}. 

With these definitions, we propose a new method to \emph{implicitly} represent the region-based homogeneity features carried by the velocity $\xi$. The basic idea is to define a set $\Theta_\fz\subset\Omega$ which contains the point $\fz$ and the initial shape $R_0$
\begin{equation}
\label{eq_Bound}
\Theta_\fz:=\{\fx\in\Theta\cup R_0;\fx\text{~is connected to~}\fz\},
\end{equation} 
where $\Theta$ consists of all the points $\fx$ such that $\xi(\fx)\leq 0$, i.e. 
\begin{equation*}
\Theta:=\{\fx\in\Omega;~\xi(\fx)\leq 0\}.
\end{equation*} 
From the definition~\eqref{eq_Bound}, one can see that the points involved in the set $\Theta_\fz$ are connected to the point $\fz$. As a result, this constraint is able to yield more selective user invention. In the following, we denote by $\partial\Theta_\fz$ the boundary of the set $\Theta_\fz$, which excludes the boundaries of holes inside $\Theta_\fz$. In other words, when traveling forward along  the boundary $\partial\Theta_\fz$ with a counter-clockwise direction, the interior of $\Theta_\fz$ is on the left.

\begin{figure*}[t]
\centering
\includegraphics[width=17.5cm]{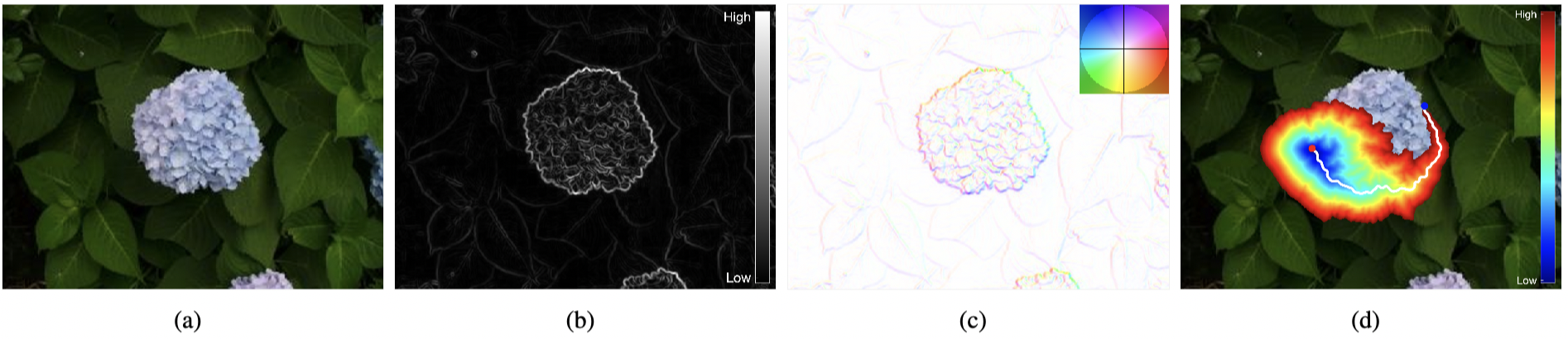}
\caption{Examples for image edge-based features. (\textbf{a}) The original image. (\textbf{b}) Visualization for the appearance features $g$. (\textbf{c}) Visualization for the vector field $\omega$ using the tool of color coding. (\textbf{d}) The geodesic distances  superimposed on the original image. The red and blue dots respectively denote the source and end points. The white line indicates the corresponding geodesic path }
\label{fig_ColorCoding}
\end{figure*}

A local minimizer for the functional $E$ in Eq.~\eqref{eq_GeneralFunctional} should satisfy the respective Euler-Lagrange equation $\partial E/\partial\cC=-\xi\cN=0$. This means that a minimizing curve should pass through the zero-level curve of the velocity $\xi$. Based on this observation, we exploit the boundary $\partial\Theta_\fz$ for the computation of the weighted function $\psi_\fz$ used in Eq.~\eqref{eq_targetMetric}. 
In our model, we expect that the geodesic paths associated to the metric $\rQ_\fz$ formulated in Eq.~\eqref{eq_targetMetric} pass through the regions close to $\partial\Theta_\fz$. This can be done by choosing $\psi_\fz$ such that it takes low values around the boundary $\partial\Theta_\fz$ and high values, otherwise. Toward this purpose, we define a Euclidean distance map $\cD:\Omega\to\bR^+_0$ with respect to the boundaries $\partial\Theta_\fz$
\begin{equation}
\label{eq_EucDist}
\cD(\fx)=\min_{\fy\in\partial\Theta_\fz}\|\fx-\fy\|.	
\end{equation}
Basically, the values of $\psi_\fz(\fx)$ should be positively correlated to the Euclidean distance value $\cD(\fx)$. In practice, one can set  $\psi_\fz(\fz)=\infty$ and for any point $\fx\in\Omega\backslash\{\fz\}$
\begin{equation}
\label{eq_DefPsi}
\psi_\fz(\fx)=f(\|\fx-\fz\|)\exp(\mu\,\cD(\fx)),	
\end{equation}
where $\mu\in\bR^+$ is a constant, and $f(a)$ is a decreasing function for $a\in\bR^+$ so as to prevent the geodesic curves from shrinking to $\fz$. As in~\cite{appleton2005globally}, we make use of $f(a)=a^{-1}$ in the experiments.

Note that the set $\Theta_\fz$ was also considered in~\cite{spencer2019parameter} to incorporate user intervention into the computation of the velocity $\xi$, where the image segmentation was implemented by a convex relaxation framework.  However, neither the asymmetric edge-based features nor the curvature regularization term were considered in~\cite{spencer2019parameter}. In contrast, these effective features can be naturally involved in the geodesic metrics considered, as introduced in the following sections.

\subsection{Computation for the Shape Gradients}
\label{Subsec_ShapeGradient}
The computation for the velocity  $\xi$ relies on the $L^1$ shape gradient of a differentiable functional $\tilde\cE:L^1(\Omega,\bR)\to\bR$. For any admissible perturbation $\delta \varrho\in L^1(\Omega,\bR)$, we can express $\tilde\cE(\varrho)$ as follows
\begin{equation}
\label{eq_LinearAppro}
\tilde\cE(\varrho+\delta \varrho)=\tilde\cE(\varrho)+\int_\Omega \delta \varrho(\fx)\xi(\fx)d\fx+o(\|\delta \varrho\|_{L^1}),
\end{equation}
where  $\xi\in L^\infty(\Omega,\bR)$ is the shape gradient of $\tilde\cE$ at $\varrho$.  It can be generated through the G\^ateaux derivative of $\tilde\cE$ 
\begin{equation}
\label{eq_ShapeGradient}
\int_\Omega\xi(\fx)f(\fx)\, d\fx=\lim_{a\to 0}\frac{\tilde\cE(\varrho+af)-\tilde\cE(\varrho)}{a}.
\end{equation}
As in~\cite{chen2019eikonal}, for a given shape $R_0$ and an arbitrary  shape  $R$ that is close to $R_0$, one can choose $\varrho=\chi_{R_0}$ and  $\delta\varrho=\chi_R-\chi_{R_0}$ such that the second term in the right-hand side of Eq.~\eqref{eq_LinearAppro} can be expressed as
\begin{equation}
\label{eq_LinearApprox2}
\int_\Omega\delta\varrho\,\xi d\fx=\int_\Omega\xi\,\chi_R d\fx-\int_\Omega\xi\,\chi_{R_0} d\fx.
\end{equation}
The first term $\int_\Omega\xi\,\chi_R d\fx$ of Eq.~\eqref{eq_LinearApprox2} is identical to the right-hand side of Eq.~\eqref{eq_GeneralFunctional}, where the velocity $\xi$ is associated to the given shape $R_0$.

We take the piecewise constants-based functional~\eqref{eq_PConst} as an example for the computation of the velocity $\xi$. In this case,  one can set $\tilde\cE(\chi_{R})=\cE(\cC)$ such that 
\begin{equation*}
\tilde\cE(\chi_{R})=\int_{\Omega}\Big(\chi_{R} \,\|I-\mathbf{c}_1\|^2+(1-\chi_R)\,\|I-\mathbf{c}_2\|^2\Big)d\fx,	
\end{equation*}
For a given shape $R_0$, the velocity $\xi(\fx)$ at $\chi_{R_0}$ can be computed by Eq.~\eqref{eq_ShapeGradient} as follows 
\begin{equation*}
\xi(\fx)= \|I(\fx)-\mathbf{c}_1\|^2-\|I(\fx)-\mathbf{c}_2\|^2,
\end{equation*}
where $\mathbf{c}_1$ and $\mathbf{c}_2$ are estimated using Eq.~\eqref{eq_Means} associated to $R_0$. 

\subsection{Metrics for Anisotropic and Asymmetric Geodesic Curves}
\label{subsec_SpatialMetric}
In geodesic paths-based image segmentation, most of existing approaches usually utilize a type of Riemannian metrics based on the edge-based features~\cite{appleton2005globally,mille2009geodesically,mille2015combination}. In order to take image edge asymmetry features into consideration, we make use of a Finsler metric with an asymmetric quadratic form~\cite{chen2018asymmetric,duits2018optimal}. Let $\bS_2^+$ be a set collecting all the positive definite symmetric matrices of size $2\times2$. Basically, the asymmetric quadratic metric $\cF:=\cF^{\rm AQ}$ can be formulated by a tensor field $\cM:\Omega\to\bS_2^+$ and  a vector field $\omega:\Omega\to\bR^2$  
\begin{equation}
\label{eq_AsyQuad}
\cF^{\rm AQ}(\fx,\fu)=\sqrt{\langle\fu,\cM(\fx)\fu \rangle+\langle\omega(\fx),\fu\rangle_+^2}\,,
\end{equation}	
where $\langle\fu_1,\fu_2\rangle_+=\max\{0,\langle\fu_1,\fu_2\rangle\}$ is the positive part of $\langle\fu_1,\fu_2\rangle$ over $\bR^2$, and $\langle\fu_1,\fu_2\rangle_+^2=(\langle\fu_1,\fu_2\rangle_+)^2$.  

The computation of the tensor field  $\cM$ and the vector field $\omega$ relies on the  the matrices $\cW(\fx)$ (see Eq.~\eqref{eq_W}), which consists of both the image edge appearance and anisotropy features. For an edge point $\fx$, the eigenvector $\vartheta(\fx)\in\bR^2$ of the matrix $\cW(\fx)$ corresponding to the smaller eigenvalue is perpendicular to the edge direction at $\fx$.  Thus, we can utilize $\vartheta(\fx)^\perp$, the perpendicular vector of $\vartheta(\fx)$, to characterize the edge anisotropy feature at $\fx$. With these definitions, the tensor field $\cM$ can be expressed as follows:
\begin{align}
\label{eq_EdgeM}
\cM(\fx)=\exp&(\alpha g(\fx))\,\vartheta(\fx)^\perp\otimes\vartheta(\fx)^\perp\nonumber\\
&+\exp(\tilde\alpha g(\fx))\vartheta(\fx)\otimes\vartheta(\fx),	
\end{align}
where  $\fu_1\otimes \fu_2=\fu_1\fu_2^T,\forall \fu_1,\fu_2\in\bR^2$. The parameters $\alpha,\,\tilde\alpha\in\bR$ (s.t. $\tilde\alpha\geq\alpha$) controls the anisotropy ratio of $\cM(\fx)$. For the sake of simplicity, we set $\tilde\alpha=0$ and $\alpha<0$ to generate anisotropic tensors $\cM(\fx)$.

In order to compute the vector field $\omega$, we first consider the gradients of the Gaussian-smoothed images, denoted by $\varpi:\Omega\to\bR^2$, expressed as 
\begin{equation}
\label{eq_asyGradient}
\varpi(\fx)=\frac{1}{3}\sum_{k=1}^3 (\nabla G_\sigma\ast I_k)(\fx).
\end{equation}
Obviously, the edge asymmetry features are carried out by the vector field $\varpi$. At an edge point $\fx$, the vector $\varpi(\fx)$ is perpendicular to the edge tangent at this point.  Thus the vector field $\omega$ can be generated using $\varpi$ as follows 
\begin{equation} 
\omega(\fx)=
\begin{cases}
\lambda\mathbf{M}\varpi(\fx)/\|\varpi(\fx)\|,&\text{if~}\|\varpi(\fx)\|\neq 0\\
\mathbf{0},&\text{otherwise},
\end{cases}
\end{equation}
where $\mathbf M$ is  a \emph{counter-clockwise} rotation matrix with rotation angle $\pi/2$ and where $\lambda\in\bR$ is a scalar parameter. Let $\tilde\cC$ be a closed curve defined over $[0,1]$ that parameterizes the target boundary in a \emph{counter-clockwise} order\footnote{In the remaining of this paper, we assume that all the closed curves are parameterized in a counter-clockwise order.}. The sign of the parameter $\lambda$ is chosen being such that the scalar products $\langle \tilde\cC^\prime(u),\omega(\tilde\cC(u))\rangle<0,~\forall u\in[0,1]$, are satisfied as much as possible.
As a consequence, the metric $\cF^{\rm AQ}$ is suitable for handling images satisfying a coherence prior on image gradients,  which can be formulated as: the scalar products $\langle\varpi(\tilde\cC(u)),\cN(u)\rangle$ along most parts of $\tilde\cC$ have the identical sign~\cite{chen2019active,kimmel2003regularized,zach2009globally}, where $\cN(u)$ is the unit normal to $\tilde\cC(u)$.

In Fig.~\ref{fig_ColorCoding}c, we illustrate an example for the visualization of the vector field $\varpi$ using the tool of color coding. The edge appearance feature map carried out by the function $g$  defined in Eq.~\eqref{eq_Magnitude} is shown in Fig.~\ref{fig_ColorCoding}b. In Fig.~\ref{fig_ColorCoding}d, we illustrate the geodesic distances associated to the metric $\cF^{\rm AQ}$, which exhibits strongly asymmetric property.

\noindent\emph{Remark}.
In many scenarios, the image segmentations can benefit from the prior on the image gradients as stated above. However, in case the prior is not satisfied, one can make use of an anisotropic Riemannian metric as a reduction of $\cF^{\rm AQ}$
\begin{equation}
\label{eq_RiemannReduction}
\cF^{\rm R}(\fx,\fu)=\sqrt{\langle\fu,\cM(\fx)\fu\rangle},
\end{equation}
by setting the vector field $\omega\equiv\mathbf{0}$.

\subsection{Metrics for Curvature-penalized Geodesic Curves}
\label{subsec_Curvature}
The curvature-regularized minimal path approaches~\cite{chen2017global,duits2018optimal,mirebeau2018fast} search for globally minimizing paths in an orientation-lifting domain $\bM=\Omega\times\bS^1$. Any point $\tilde\fx=(\fx,\theta)\in \Omega\times\bS^1$ is made up of a physical position $\fx\in\Omega$ and an angular coordinate $\theta\in\bS^1$.
In the proposed model, the component $\cF$ of the metric $\rQ_\fz$ in Eq.~\eqref{eq_targetMetric} can be a metric with curvature regularization. Examples for curvature-penalized geodesic approaches may involve the Euler-Mumford elastica geodesic model~\cite{chen2017global} and the Reeds–Shepp forward model~\cite{duits2018optimal}. The key idea for both models is to represent the tangent directions of a smooth planar curve $\gamma:[0,1]\to\Omega$ via an orientation lifting $\varphi:[0,1]\to\bS^1$ such that for any $u\in[0,1]$ 
\begin{equation}
\label{eq_OnWay}
\gamma^\prime(u)=\|\gamma^\prime(u)\|\fn(\varphi(u)),
\end{equation}
where $\fn(\theta)=(\cos\theta,\sin\theta)^T$.
Simple calculation yields that the curvature $\kappa:[0,1]\to\bR$ of a curve $\gamma$ can be denoted by the ratio of $\varphi^\prime$ and $\|\gamma^\prime\|$, i.e. $\kappa=\varphi^\prime/\|\gamma^\prime\|$. 

The weighted curve length involving the curvature $\kappa$ along a curve $\tilde\gamma=(\gamma,\varphi):[0,1]\to\Omega\times\bS^1$ satisfying~\eqref{eq_OnWay} can be formulated as
\begin{align}
\label{eq_MetricInterpretation2}
\cL(\tilde\gamma)&=\int_0^1 \cP(\tilde\gamma(u))\,\big(1+\beta\kappa(u)^2\big)^\varsigma \|\gamma^\prime(u)\|\,du\\
&=\int_0^1 \cP(\tilde\gamma(u))\,\left(1+\frac{\beta\varphi^\prime(u)^2}{\|\gamma^\prime(u)\|^2}\right)^\varsigma \|\gamma^\prime(u)\|\,du\nonumber\\
&=\int_0^1\cF(\tilde\gamma(u),\tilde\gamma^\prime(u))\,du,
\label{eq_MetricInterpretation}
\end{align}
where $\cP:\Omega\times\bS^1\to\bR^+$ is an orientation-dependent function and $\beta\in\bR^+$ is a parameter that weights the importance of the curvature. The metric $\cF$ can be expressed for any point $\tilde\fx=(\fx,\theta)\in\Omega\times\bS^1$ and any vector $\tilde\fu=(\fu,\nu)\in\bR^3$
\begin{equation*}
\cF(\tilde\fx,\tilde\fu)=
\begin{cases}
\cP(\tilde\fx)\left(1+\frac{\beta\nu^2}{\|\fu\|^2}\right)^\varsigma\|\fu\|,&\text{if~} \fu=\fn(\theta)\|\fu\|,\\
\infty,&\text{otherwise}.
\end{cases}
\end{equation*}
The metric $\cF$ used in Eq.~\eqref{eq_MetricInterpretation} with $\varsigma=1$ and $\varsigma=1/2$ respectively corresponds to the Euler-Mumford elastica metric~\cite{chen2017global} and the Reeds-Shepp forward metric~\cite{duits2018optimal}. 
Finally, the data-driven function $\cP$ can be defined as~\cite{chen2017global}
\begin{equation}
\label{eq_OrienPotential}
\cP(\fx,\theta)=\exp\left(\alpha\,\langle\fn(\theta)^\perp,\cW(\fx)\fn(\theta)^\perp\rangle\right),
\end{equation}
where $\alpha<0$ is a scalar-valued parameter. In this definition, the term $\langle\fn(\theta)^\perp,\cW(\fx)\fn(\theta)^\perp$ stands for the orientation score. One can point out that if the vector $\fn(\theta)^\perp$ is proportional to the edge tangents for edge points $\fx$, the values of $\cP(\fx,\theta)$ are low, satisfying the requirement in image segmentation applications.

\begin{figure*}[t]
\centering
\includegraphics[width=17.5cm]{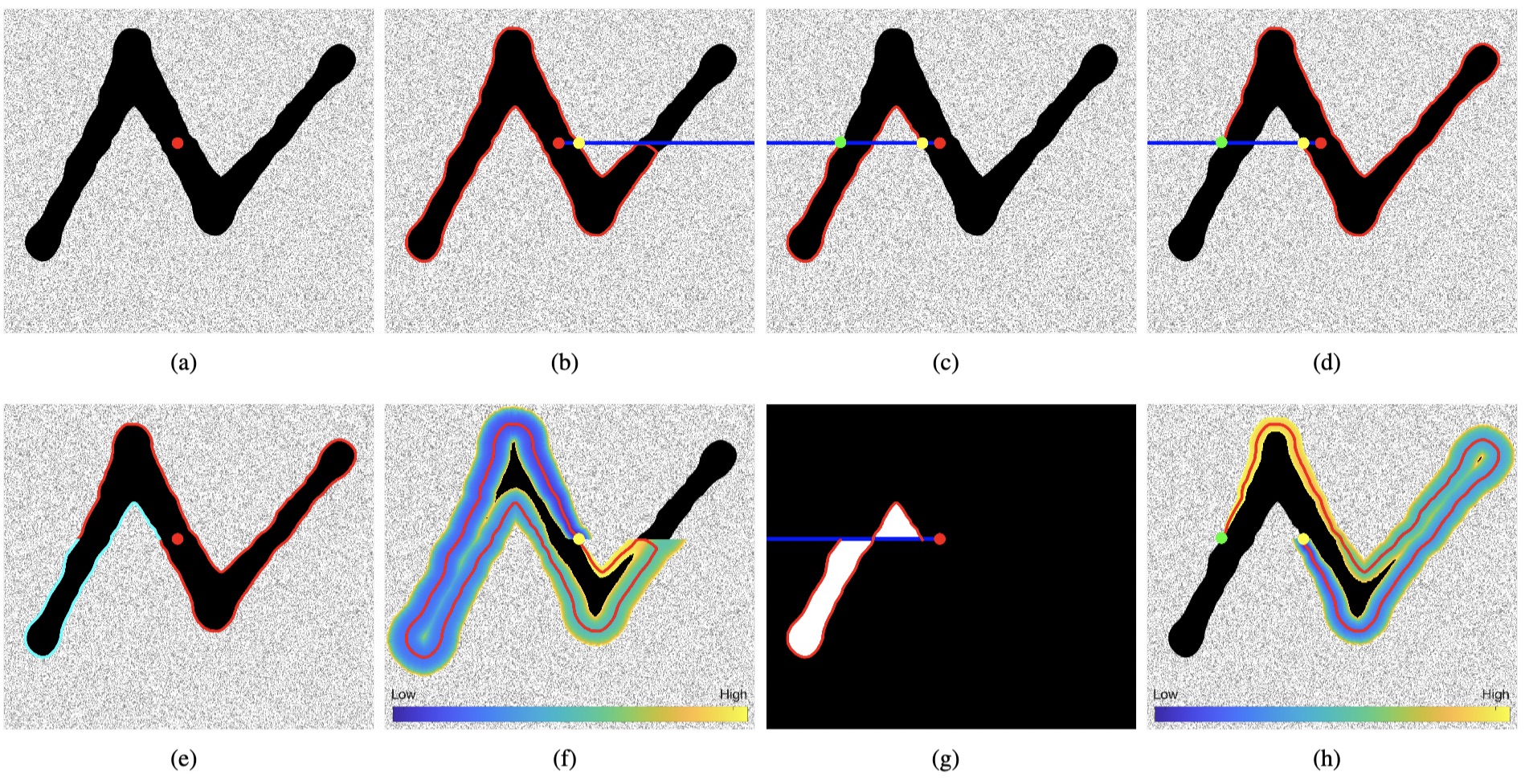}
\caption{Illustration for the proposed dual-cut scheme. (\textbf{a}) A synthetic image with a red dot denoting the landmark point $\fz$. (\textbf{b}) The blue solid line indicates $\ell_\fz^+$, the red line is the geodesic path $\cG_{\fq}$ obtained in the first step and the yellow dot indicates the sampled point $\fq$. (\textbf{c}) The green and yellow dots are the points $\fa$ and $\fb$. The red line is the  $\Gamma_{\fa,\fb}$ which is a portion of $\cG_{\fq}$ and the blue line denotes $\ell^-_\fz$. 
(\textbf{d}) The geodesic path $\cG_{\fb,\fa}$ denoted by red line. (\textbf{e}) The target closed curve $\cC$, see text. (\textbf{f}) and (\textbf{h}) The geodesic distance maps superimposed on the original image, which correspond to figures (b) and (d). (\textbf{g}) The region $\Theta$ tagged as white color
}
\label{fig_Dualcut}
\end{figure*}

\section{Dual-cut Scheme for Closed Curve Detection}
\label{sec_DualCut}

Finding image segmentation under a geodesic framework usually amounts to building simple and closed curves made up of geodesic paths. In this section, we introduce a dual-cut scheme, which is an adaption of the circular geodesic model~\cite{appleton2005globally}, to efficiently solve the interactive image segmentation problems. 

In the circular geodesic model, the basic idea is to impose a constraint to the image domain $\Omega$ by means of a cut, where the origin of the domain $\Omega$ is instantiated in the given landmark  point $\fz$, and the cut is placed infinitesimally beneath the non-negative $x$-axis.  Hereinafter we denote by $\ell^+_\fz$ the non-negative $x$-axis. The use of the cut adds disconnection constraint between the two sides of the cut,  which favours to detect \emph{cut-convexity} curves~\cite{appleton2005globally}. For convenience, we denote by $\Xi_\fz$ the set collecting all the cut-convexity curves with respect to the given landmark point $\fz$ such that
\begin{align}
\Xi_\fz:=\Big\{\gamma\in\Lip([0,1],\Omega);~&\fz\in R_\gamma,\,\gamma(0)=\gamma(1)\in\ell^+_\fz,\nonumber\\
&\gamma(u)\notin\ell^+_\fz,\forall u\in(0,1)\Big\},
\end{align}
where $R_\gamma$ represents the interior region of closed curve $\gamma$. We emphasize that the  point $\fz$ is not passed by $\gamma$.

\subsection{Initialization}
\label{subsec_Initializations}
The generation of an initial shape $R_{0}\subset\Omega$ such that  $\fz\in R_0$ is the first stage of the proposed method. Basically, this shape $R_0$ is expected to be covered by the target region as much as possible. Towards this purpose, we choose to  construct $R_0$ by means of a  front propagation procedure expanding from $\fz$. Specifically, this can be done by thresholding a geodesic distance map which admits the solution to the following isotropic Eikonal PDE
\begin{equation}
\label{eq_InitializationPDE}
\|\nabla\cU_{\fz}(\fx)\|=\phi(\fx),\quad\forall\fx\in\Omega\backslash\{\fz\},
\end{equation}
with boundary condition $\cU_{\fz}(\fz)=0$. The function $\phi:\Omega\to\bR^+$ is an edge indicator defined by 
\begin{equation}
\phi(\fx)=\exp(\tau\,g(\fx))-\tau_\epsilon,
\end{equation}
where $\tau\in\bR^+$ and $\tau_\epsilon\in(0,1)$ are two constants, and $g$ is the magnitude of image gradients as defined in Eq.~\eqref{eq_Magnitude}. In our experiments, we fix $\tau=5$ and $\tau_\epsilon=0.99$ for all the numerical experiments. 
Then, the  shape $R_{0}$ can be generated by 
\begin{equation}
\label{eq_InitialShape}
R_0=\{\fx\in\Omega;\,\cU_{\fz}(\fx)\leq T\},
\end{equation}
where $T\in \bR^+$ is a thresholding value.
Once the construction of $R_0$ is done,  we can estimate the shape gradient $\xi$ with respect to $R_{0}$ using Eq.~\eqref{eq_ShapeGradient}.   

\subsection{Dual-cut Scheme for Closed Contour Detection}
\label{subsec_Principle}

Suppose that the set $\Theta_\fz$ has been built by means of Eq.~\eqref{eq_Bound}. Basically, the proposed dual-cut scheme can be divided into two steps, where the target is to seek a simple closed curve as the concatenation of two disjoint geodesic paths. 

\subsubsection{Step I}
In the first step, among all the intersection points between the non-negative $x$-axis $\ell_\fz^+$ and the boundary $\partial\Theta_\fz$, we  choose a point $\fq\in\ell_\fz^+$ that is closest to $\fz$ in the sense of Euclidean distance. The goal in this step is to extract a geodesic curve from the set $\Xi_\fz$.  Such a geodesic path can be generated by solving the following minimizing problem
\begin{equation}
\label{eq_CutGeo}
\cG_{\fq}=\underset{\gamma\in\Xi_\fz}{\arg\min}\left\{\int_0^1 \rQ_\fz(\gamma,\gamma^\prime)du;\gamma(0)=\gamma(1)=\fq\right\}.
\end{equation}
We give an example in Fig.~\ref{fig_Dualcut}b for this step using a synthetic image. In this figure, the red and yellow dots respectively indicate the points $\fz$ and $\fq$, and  the cyan line represents $\ell^+_\fz$. By the definition~\eqref{eq_CutGeo}, the closed geodesic curve $\cG_{\fq}$ is allowed to pass through the non-negative $x-$axis only once. As a result, $\cG_{\fq}$ may fail to delineate the boundary segments which pass through  $\ell^+_\fz$ multiple times. This can be seen from Fig.~\ref{fig_Dualcut}b, where the boundary segment at the top right corner is missed.  In order to overcome this issue, we consider to use the non-positive $x$-axis, referred to as $\ell_\fz^-$, to tack another geodesic path in order to delineate the boundary segments missed by $\cG_\fq$.

\subsubsection{Step II}
\label{subsubsec_StepII}
From the first step, we have obtained a simple closed geodesic curve $\cG_{\fq}$, which intersects with non-positive axis $\ell_\fz^-$ at least once. Among these intersection points, we choose the first and the last points, respectively denoted by $\cG_\fq(u_1)=\fa\in\ell^-_\fz$ and $\cG_\fq(u_2)=\fb\in\ell^-_\fz$ such that $0<u_1\leq u_2<1$, in order to track a new geodesic curve.

\noindent\emph{In case $\fa\neq\fb$}. We denote by $\cG_\fq|_{u_1\to u_2}$ the portion of the geodesic path $\cG_{\fq}$ traveling from $\fa$ to $\fb$, and denote by $\Gamma_{\fa,\fb}$ the re-parameterization of $\cG_\fq|_{u_1\to u_2}$ over the range $[0,1]$, i.e. 
\begin{equation}
\Gamma_{\fa,\fb}(0)=\fa,\quad\Gamma_{\fa,\fb}(1)=\fb.
\end{equation}
An example for the path $\Gamma_{\fa,\fb}$ can be seen from Fig.~\ref{fig_Dualcut}c,  indicated by a red line. 

Let $A\subset\Omega$ be a set which is regarded as the union of closed regions enclosed by the non-positive $x$-axis $\ell_\fz^-$ and $\Gamma_{\fa,\fb}$, see Fig.~\ref{fig_Dualcut}g for an example. We expect to seek an open geodesic path $\cG_{\fb,\fa}$,  which links $\fb$ to $\fa$ and is forbidden to pass through $A$. For this purpose, we consider a set of curves
\begin{align*}
\Upsilon_{\fz}=\Big\{\gamma\in\Lip([0,1],\Omega);~&\gamma(0)\in\ell_\fz^-,~\gamma(1)\in\ell_\fz^-,\\
&\gamma(u)\notin\ell^-_\fz\cup A,\forall u\in(0,1)\Big\}
\end{align*}
yielding that 
\begin{equation}
\label{eq_CutGeo2}
\cG_{\fb,\fa}=\underset{\gamma\in\Upsilon_\fz}{\arg\min}\left\{\int_0^1 \rQ_\fz(\gamma,\gamma^\prime)du\right\}s.t.\,
\begin{cases}
\gamma(0)=\fb,\\
\gamma(1)=\fa.	
\end{cases}
\end{equation}
We illustrate an example for such a geodesic path $\cG_{\fb,\fa}$ in Fig.~\ref{fig_Dualcut}d. 
Similar to the classical geodesic tracking procedure as introduced in Section~\ref{subsec_MP},  the minimization of the problems~\eqref{eq_CutGeo} and~\eqref{eq_CutGeo2} can be addressed by estimating  geodesic distance maps associated to the respective metrics $\rQ_\fz$, implemented via a variant of the fast marching methods~\cite{mirebeau2018fast,mirebeau2019riemannian} in conjunction with $\ell^+_\fz$ and $\ell^-_\fz$, see Section~\ref{subsec_NI}.

Now we can build the target closed curve $\cC\in\Lip([0,1],\Omega)$ as the concatenation of two paths $\Gamma_{\fa,\fb}$ and $\cG_{\fb,\fa}$, i.e. 
\begin{equation}
\label{eq_TargetCurve}
\cC(u)=\left(\Gamma_{\fa,\fb}\doublecup\cG_{\fb,\fa}\right)(u),\quad \forall u\in[0,1]	
\end{equation}
where $\doublecup$ is a concatenation operator of two curves $\gamma_1$ and $\gamma_2:[0,1]\to\Omega$
\begin{equation}
\label{eq_Concatenation}
(\gamma_1\doublecup\gamma_2)(u)=
\begin{cases}
\gamma_1(u),&\text{if~} u\in[0,1/2],\\
\gamma_2(u),&\text{if~} u\in[1/2,1].
\end{cases}
\end{equation}

\begin{figure*}[t]
\centering
\includegraphics[width=17.5cm]{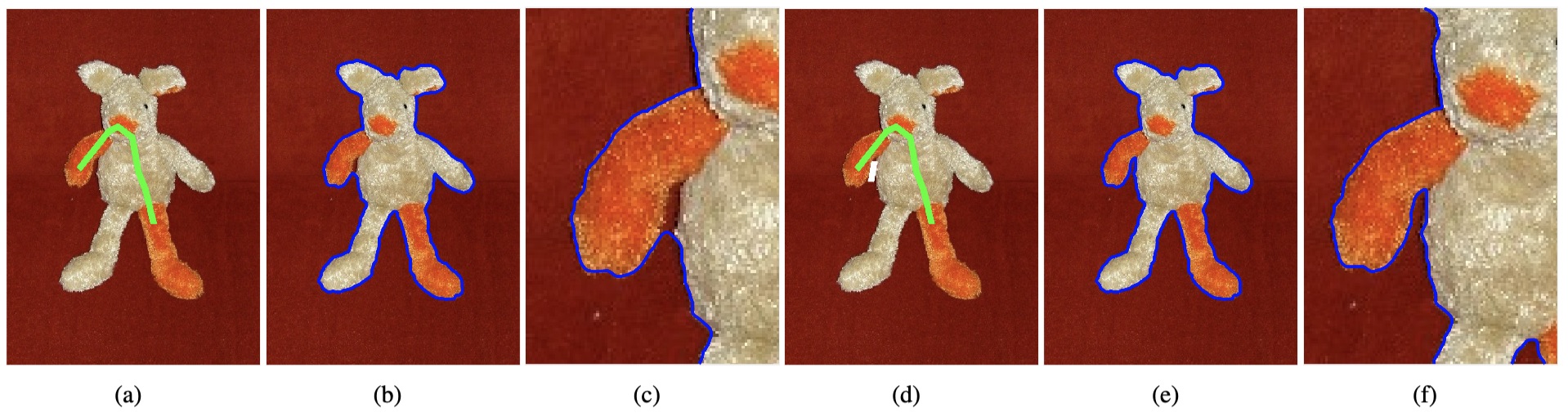}
\caption{Image segmentation derived from the proposed model using scribbles. (\textbf{a}) User-provided scribbles indicated by a green line. \textbf{b} Image segmentation contour indicated by a blue line. (\textbf{c}) Close-up view of the segmentation contour in Figure~(b). (\textbf{d}) An additional scribble (white line) used as a barricade. (\textbf{e}) Image segmentation contour constrained by all the scribbles. (\textbf{f}) Close-up view of the segmentation contour in Figure (e)}
\label{fig_Scribble}
\end{figure*}

\noindent\emph{In case $\fa=\fb$}.
In particular, if the path $\cG_{\fq}$ crosses over the non-positive $x$-axis $\ell^-_\fz$ just once, then we have $\fa=\fb$. In this case, the closed set $A=\emptyset$ and the geodesic paths $\cG_{\fb,\fa}$ can be still defined by~\eqref{eq_CutGeo2}. In this case, the concatenation operator is no longer needed such that the target curve $\cC=\cG_{\fb,\fa}$.

\noindent\emph{Using curvature-penalized geodesic paths}.
In this section, the geodesic paths are assumed to be defined over the image domain $\bM=\Omega$, which are suitable for the case of asymmetric quadratic metrics $\cF^{\rm AQ}$. With respect to the curvature-penalized geodesic metrics over the domain $\bM:=\Omega\times\bS^1$, the minimization problems~\eqref{eq_CutGeo} and~\eqref{eq_CutGeo2} respectively get to be 
\begin{equation}
\label{eq_CutGeo2}
\tilde\cG_{\fq}=\mathop{\arg\min}_{\scriptstyle\tilde\gamma=(\gamma,\varphi)\in\Lip([0,1],\bM)
\atop
\scriptstyle\tilde\gamma(0)=\tilde\gamma(1)=\tilde\fq
}
\left\{\int_0^1 \rQ_\fz(\tilde\gamma,\tilde\gamma^\prime)du;\gamma\in\Xi_\fz\right\}
\end{equation}
and 
\begin{equation}
\label{eq_CutGeo3}
\tilde\cG_{\fb,\fa}=\mathop{\arg\min}_{\scriptstyle\tilde\gamma=(\gamma,\varphi)\in \Lip([0,1],\bM)
\atop
\scriptstyle\tilde\gamma(0)=\tilde\fb,\tilde\gamma(1)=\tilde\fa
}\left\{\int_0^1 \rQ_\fz(\tilde\gamma,\tilde\gamma^\prime)du;\gamma\in\Upsilon_{\fz}\right\}.
\end{equation}

When applying the curvature-penalized geodesic paths for the proposed dual-cut scheme, the sampled point $\fq$, used in the first step, is lifted to $\tilde\fq=(\fq,\theta_\fq)$ such that 
\begin{equation*}
\theta_\fq=\underset{\theta\in(0,\pi)}{\arg\min}\,\cP(\fq,\theta).	
\end{equation*}
In the second step, the intersection points $\fa$ and $\fb$ are detected using the physical projection $\cG_\fq$ of the orientation-lifted geodesic path $\tilde\cG_\fq=(\cG_\fq,\varphi_\fq)$, see Eq.~\eqref{eq_CutGeo2}. Again, in order to generate the orientation-lifted geodesic path $\tilde\cG_{\fb,\fa}$, one should respectively lift $\fa$ and $\fb$ to $\tilde\fa=(\fa,\theta_\fa)$ and $\tilde\fb=(\fb,\theta_\fb)$
\begin{equation*}
\theta_\fa=\underset{\theta\in(-\pi,0)}{\arg\min}\,\cP(\fa,\theta),\quad \theta_\fb=\underset{\theta\in(-\pi,0)}{\arg\min}\,\cP(\fb,\theta).
\end{equation*}
Note that the detection of the orientations $\theta_\fq,\,\theta_\fa$ and $\theta_\fb$  coincides with the assumption that the curves $\cG_\fq$ and $\cC$ are parameterized in a counter-clockwise order.

\begin{figure*}[t]
\centering
\includegraphics[width=17.5cm]{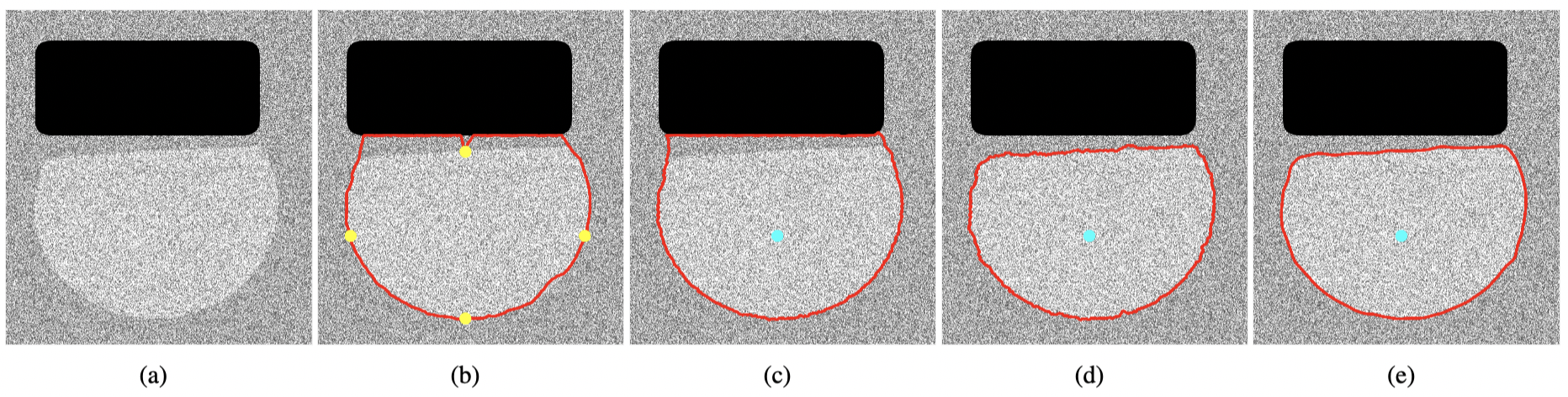}
\caption{Influence of the implicit region-based homogeneity features on the final segmentation contours. (\textbf{b}) and (\textbf{c}) Segmentations from the CombPaths  and VCGeo models, respectively. (\textbf{d}) and (\textbf{e}) Segmentation results derived from the proposed DualCut-Asy and DualCut-RSF models, respectively}
\label{fig_syn}
\end{figure*}
\begin{figure*}[t]
\centering
\includegraphics[width=17cm]{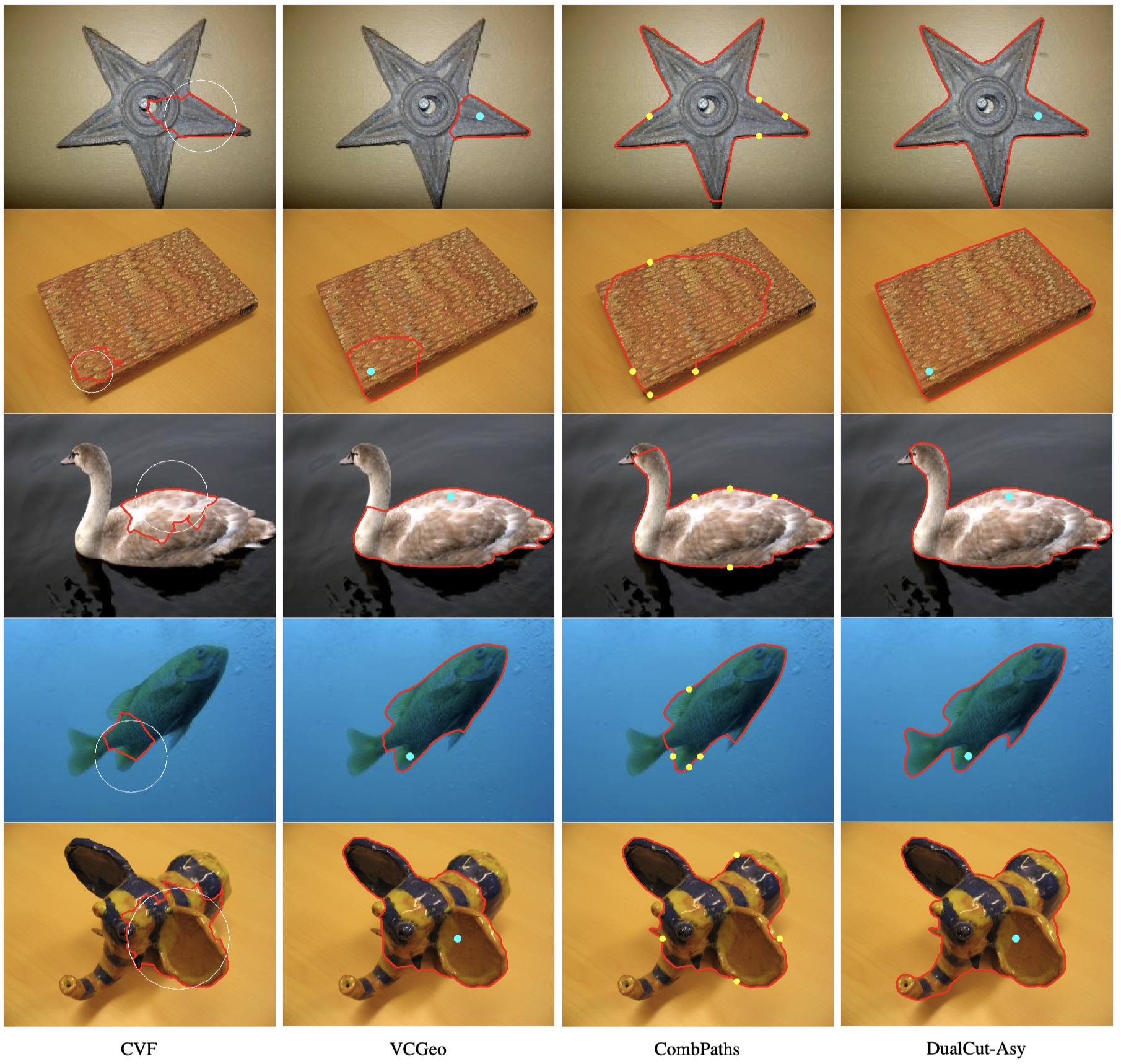}
\caption{Qualitative comparison results with the CVF model, the VCGeo model and the CombPaths model. The segmentation contours are indicated by red lines.  The while in column $1$ are initial contours for the CVF model.  The cyan dots in columns $2$ and $4$ indicate the user-provided points for the VCGeo and DualCut-Asy models. The yellow dots are the input points for the CombPaths model}
\label{fig_examples_natural}	
\end{figure*}

\subsection{Extending User Intervention from a Point to Scribbles}
\label{subsec:Scribbles}
In the dual-cut closed curve detection scheme as introduced in Section~\ref{subsec_Principle}, the user input is supposed to be a single point used to locate the target region. As in many interactive image segmentation approaches~\cite{boykov2006grapy,grady2006random,chen2018asymmetric}, scribbles often serve as seeds to provide constraint for image segmentation. In this section, we present a method to add user-provided scribbles to the proposed dual-cut model.

In the basic setting of the proposed model, a scribble $\cS$ can be modeled as a continuous curve line placed in the image domain $\Omega$. The scribbles considered can be classified into two categories. The first one serves as foreground seeds to locate the regions we attempt to search for (see Fig.~\ref{fig_Scribble}a for an example), from which  one can sample the landmark  point $\fz\in\cS$. In addition, the target curves $\cG$ defined in Eq.~\eqref{eq_TargetCurve} are supposed to surround the scribbles $\cS$. In practice, the scribbles $\cS$ are used to stop the fast marching fronts to pass through it. The initial shape $R_0$ can be constructed by Eq.~\eqref{eq_InitialShape}, where the associated geodesic distance map $\cU_\cS$ satisfies the isotropic Eikonal PDE with respect to $\phi$, as used in Eq.~\eqref{eq_InitializationPDE}.

In some complicated scenarios, favorable segmentations may require more user intervention, in addition to the scribbles tagged as foreground. In Fig.~\ref{fig_Scribble}b, one can see that the segmentation contour can accurately delineate most of the target boundaries except for a high concave part, where the close-up view for this part is depicted in Fig.~\ref{fig_Scribble}c. In order to overcome this problem, we consider the second type of scribbles which serves as barricades to prevent the geodesic paths from crossing them. Furthermore, no label is assigned to the second type of scribbles. We illustrate the segmentation result using the second type of scribbles in Figs.~\ref{fig_Scribble}e and~\ref{fig_Scribble}f.

\begin{algorithm}[!t]
\caption{\textsc{Fast Marching Method}}	
\label{algo_FM}
\begin{algorithmic}
\renewcommand{\algorithmicrequire}{\textbf{Input:}}
\renewcommand{\algorithmicensure}{\textbf{Output:}}
\Require  A source point $\fs$ and a set $\Im_{\rm end}$;
\Ensure Geodesic distance map $\cU_\fs$;
\renewcommand{\algorithmicrequire}{\textbf{Initialization}:}
\Require 
\State $\bullet$ Set $\cU_\fs(\fs)\gets0$ and set $\cU_\fs(\fx)\gets\infty$, $\forall \fx\in\bZ^n\backslash\{\fs\}$.
\State $\bullet$ Set $\cV(\fx)\gets\Trial$, $\forall \fx\in\bZ^n$.
\State $\bullet$ Set $\fx_{\rm min}\gets\fs$ and $\StoppingFlag$ $\gets$ FALSE.
\end{algorithmic}
\begin{algorithmic}[1]
\renewcommand{\algorithmicrequire}{\textbf{Main Loops}:}
\Require
\While{$\StoppingFlag$ $\neq$ TRUE} 
\State Find $\fx_{\rm min}$  minimizing $\cU_\fs$ among all $\Trial$ points;
\State $\cV(\fx_{\rm min})\gets\Accepted$;
\If{$\fx_{\rm min}\in\Im_{\rm end}$}
\State Set $\StoppingFlag$ $\gets$ TRUE.
\EndIf 
\For{all $\fx_{\rm n}$ s.t. $\cV(\fx_{\rm n})= \Trial$ and $\fx_{\rm min}\in\Lambda(\fx_{\rm n})$}
\If{$\Satisfactory(\fx_{\rm min},\fx_{\rm n})=$TRUE}
\label{algLine_Satisf}
\State Update the value $\cU_\fs\fx_{\rm n})$ by solving the upwind discretization of the Eikonal PDE~\eqref{eq_HamiltionHJB}.
\label{algLine_Update}
\Else
\State $\cU_\fs(\fx_{\rm n})\gets+\infty$;
\EndIf
\EndFor
\EndWhile
\end{algorithmic}
\end{algorithm}

\subsection{Numerical Implementation}
\label{subsec_NI}
In Section~\ref{subsec_Principle}, we have introduced the dual-cut scheme for extracting closed curves. A crucial ingredient is to track two geodesic paths using the sets $\Xi_\fz$ and $\Upsilon_{\fz}$. In this section, we show that these geodesic paths can be efficiently generated by an adaption of state-of-the-art Hamiltonian Fast Marching method (HFM). We first consider the numerical implementation in the 2D domain, which can be simply extended to the orientation-lifted case.

The HFM method is based on the reformulation of the Eikonal equation~\eqref{eq_FinslerEikonal}
\begin{equation}
\label{eq_HamiltionHJB}
\cH(\fx,\nabla\cU_\fs(\fx))=\frac{1}{2},~\forall \fx\in\Omega\backslash\{\fs\},\\
\end{equation}
with $\cU_\fs(\fs)=0$ as the boundary condition, where $\cH$ is the Hamiltonian with respect to the metric $\cF$, defined as
\begin{equation*}
\cH(\fx,\fv) = \sup_{\fu\in \bR^2}\left\{\langle\fu,\fv\rangle - \frac{1}{2} \cF(\fx,\fu)^2\right\}.
\end{equation*}

The first stage for tracking the geodesic paths $\cG_\fq$ and $\cG_{\fb,\fa}$ respectively defined in Eqs.~\eqref{eq_CutGeo} and~\eqref{eq_CutGeo2} is implemented by estimating geodesic distance maps. In the HFM method, the estimation of these distance maps is performed in a regular grid $\bM_h$ with $h$ being the discretization scale. The update of distance values is implemented by solving a finite differences discretization of the Eikonal equation~\eqref{eq_HamiltionHJB},  based on the adaptive stencils $\Lambda$. For each grid point $\fx\in\bM_h$, the stencil  $\Lambda(\fx)$ is made up of a finite number of offsets $\fe_j\in\mathbb Z^2$ with integer coordinates, which are constructed using a discrete geometry tool of Voronoi’s first reduction of quadratic forms~\cite{mirebeau2018fast,mirebeau2019riemannian}. 

The constraints used to define the sets $\Xi_\fz$ and $\Upsilon_{\fz}$ are respectively introduced by the axes $\ell_\fz^+$ and  $\ell_\fz^-$, which can be incorporated into the HFM by removing unsatisfactory offsets from some stencils $\Lambda(\fx)$. Denote by $\fy_j=\fx+h\fe_j\in\bM_h,\,\forall\fe_j\in\Lambda(\fx)$ the neighbourhood points of a grid point $\fx\in\bM_h$. When computing the geodesic curve $\cG_\fq$ in the first step (resp. the geodesic curve $\cG_{\fb,\fa}$ in the second step), we consider the following two conditions: 
\begin{itemize}
	\item[(i)] If $\fx\notin \ell^+_\fz\cap\bM_h$ (resp. $\fx\notin \ell^-_\fz$) and the segment $[\fx,\fy_i]$ intersects with $\ell^+_\fz$ (resp. with $\ell^-_\fz$).
	\item[(ii)] If $\fx\in \ell^+_\fz\cap\bM_h$ (resp. $\fx\in \ell^-_\fz$) and the scalar product $\langle\fy_j-\fx,(0,1)^T\rangle<0$ (resp. $\langle\fy_j-\fx,(0,-1)^T\rangle<0$).
\end{itemize}
The offsets $\fe_j$ will be removed from the stencil $\Lambda(\fx)$ if they satisfy either condition.

Specifically, in the first step of the dual-cut scheme, we estimate a geodesic distance map $\cU_\fq$, which satisfies the Eikonal PDE~\eqref{eq_HamiltionHJB}. This is done by the HFM using the constrains related to $\ell^+_\fz$, as formulated in Points (i) and (ii).  While in the second step, likewise the distance map $\cU_\fq$, we estimate a geodesic distance map $\cU_\fb$ by the HFM with stencils following the constraints as stated in Points (i) and (ii) with respect to $\ell^-_\fz$.  In this step, we further take into account the set $A$ in addition to the conditions above,  such that an offset $\fe_j$ should be eliminated from the stencil $\Lambda(\fx)$ if the grid point $\fx+h\fe_j\in A$. 
Based on the geodesic distance maps $\cU_\fq$ and $\cU_\fb$, one can respectively obtain the geodesic paths $\cG_\fq$ and $\cG_{\fb,\fa}$,  by solving the gradient descent ODEs~\eqref{eq_ODE}. We illustrate the geodesic distance maps superimposed on the original image in Figs.~\ref{fig_Dualcut}f and~\ref{fig_Dualcut}h, which are respectively generated in the first and second step of the proposed dual-cut scheme. 
In order to reduce the computation time, the HFM will be terminated once an end point $\fx\in\Im_{\rm end}$ is reached by the fast marching fronts. In the first step of the proposed dual-cut model, the set $\Im_{\rm end}:=\{\fy\in\bM_h;\|\fy-\fq\|\leq\sqrt{2}h,\langle \fy-\fq,(0,-1)^T\rangle>0\}$. While in the second step, we use $\Im_{\rm end}:=\{\fy\in\bM_h;\|\fy-\fa\|\leq\sqrt{2}h,\langle \fy-\fa,(0,1)^T\rangle>0\}$. 
The main algorithm for estimating geodesic distance maps can be seen in Algorithm~\ref{algo_FM}. In this algorithm, the value of  $\Satisfactory(\fx,\fy_j)$ is false if the offset $\fe_j$ should be removed from the stencil $\Lambda(\fx)$. 

Finally, with respect to the curvature-penalized metrics, each offset $\tilde\fe_j=(\fe_j,e_j^\theta)\in \mathbb Z^3$  yields a neighbourhood point $\tilde\fy_j=(\fy_j,\theta_j)=\tilde\fx+h\tilde\fe_j$. In this case, Points~(i) and (ii) are checked using $\fe_j$ and the test associated to the set $A$ is examined using the physical position $\fy_j$.

\begin{figure}[t]
\centering
\subfigure[]{\includegraphics[width=4cm]{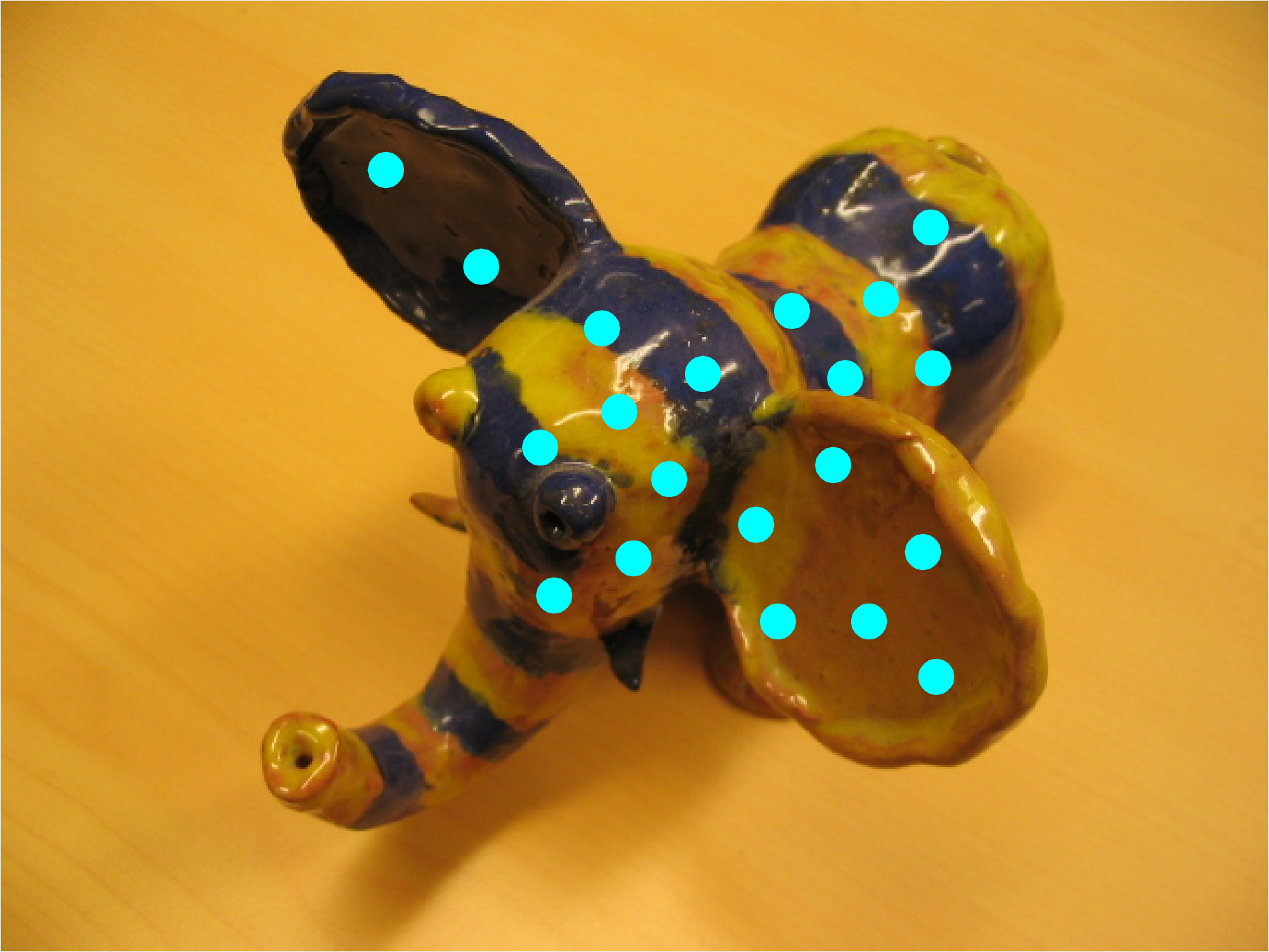}}\,\subfigure[]{\includegraphics[width=4cm]{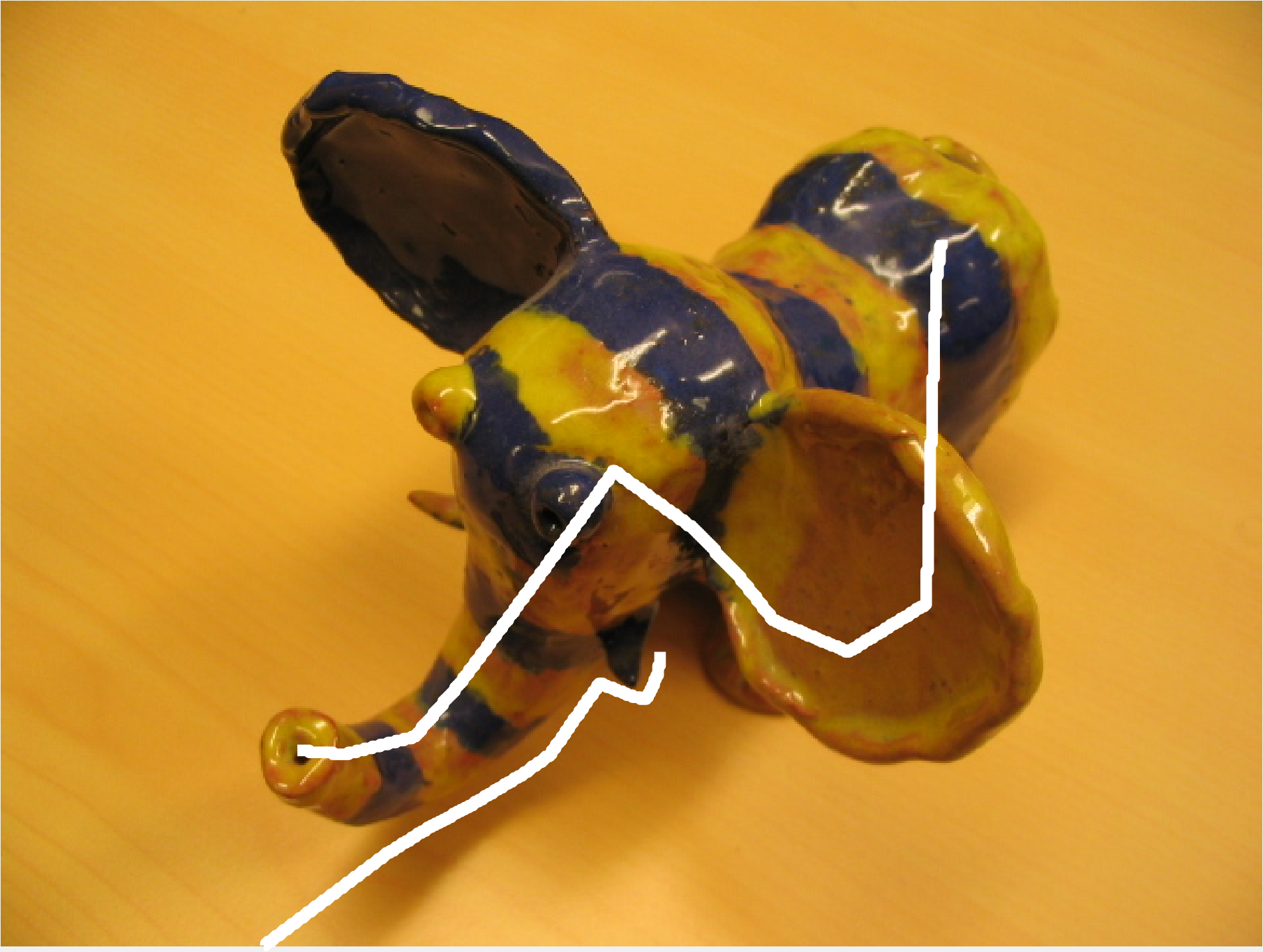}}\\
\vspace{2mm}
\subfigure[]{\includegraphics[width=8cm]{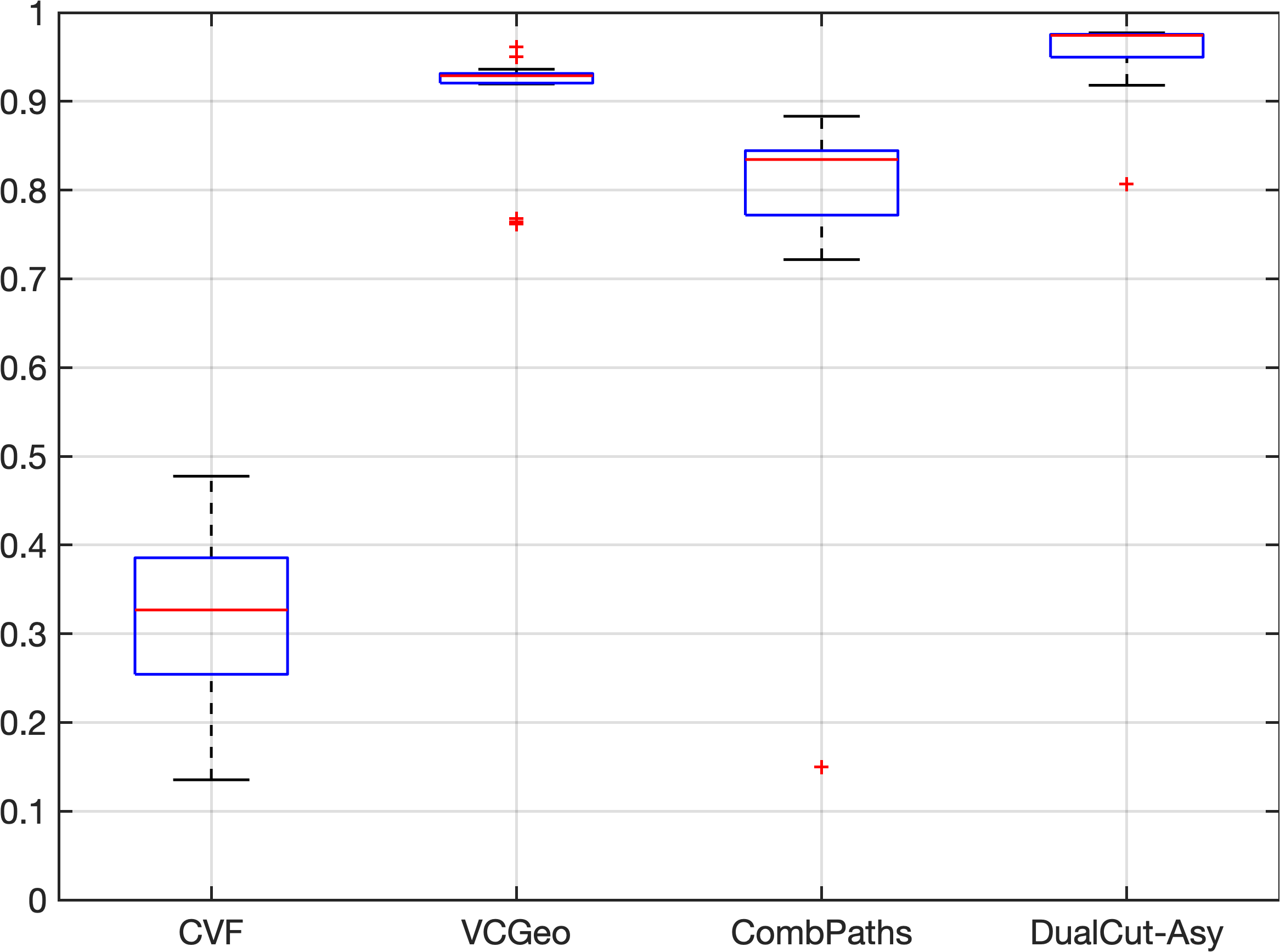}}
\caption{(\textbf{a}) The blue dots indicate a set of sampled points, each of which serves as the landmark  point $\fz$. (\textbf{b}) The scribbles of the second type which serve as obstacles. (\textbf{c}) Box plots of the Jaccard score values associated to the points shown in figure (a) for different models}
\label{fig_examples_obstacle}	
\end{figure}

\section{Experimental Results}
\label{sec_Exp}
In this section, we conduct the qualitative and quantitive comparison experiments with the vector field convolution (VFC) active contour model~\cite{li2007active}, a variant of the circular geodesic model (VCGeo), and the combination of paths (CombPaths) model~\cite{mille2015combination}. The brief introduction for the VCGeo model is presented in Appendix~\ref{appendix_VCGeo}. Finally, we use the abbreviation of  DualCut-Asy (resp. DualCut-RSF) to represent the proposed dual-cut model with a spatial asymmetric quadratic metric (resp. Reeds-Shepp Forward metric).

In this section, the quantitative evaluation is carried out by  the Jaccard score, which measures the overlap between a segmented region $\mathbb{S}$ and the ground truth region $GT$ 
\begin{equation*}
J(\mathbb{S},GT)=\frac{\#|\mathbb{S}\cap GT|}{\#|\mathbb{S}\cup GT|}	
\end{equation*}
where $\#|\mathbb{S}|$ denotes the number of grid points involved in $\mathbb{S}$.

\subsection{Parameter setting}
In the proposed  dual-cut model, the considered metrics $\rQ_\fz$, as defined in Eqs.~\eqref{eq_targetMetric},  are comprised of two components: the scalar-valued function $\psi_\fz$ that implicitly encodes the region-based homogeneity features and the Finsler metrics $\cF$. The parameter $\mu$ for $\psi_\fz$  controls the importance of the regional information. If the region-based homogeneity terms used are suitable for the image data, one can assign large values to $\mu$ and small values, otherwise. Typically, we found that $\mu=0.1$ or $\mu=0.2$ can generate favorable segmentation results.  For an asymmetric quadratic metric $\cF^{\rm AQ}$ defined in Eq.~\eqref{eq_AsyQuad}, we should build the tensor field $\cM$ and the vector field $\omega$ relying on the parameters $\alpha$ and $\lambda$, respectively.  The values of $\alpha$ dominate the importance of the edge-based features and we found that satisfactory segmentations can be obtained for $\alpha\in[5,8]$. We invoke $\alpha=7$ and $|\lambda|=2$ unless otherwise specified. Note that the values  $\sign(\lambda)$ are set in terms of the image gradients coherence prior and depend on the image data.
In the following experiments, we take the Reeds-Sheep forward metric as the instance of the curvature-penalized metrics. In this case, 
 the value of $\alpha=5$ is used for computing the orientation-dependent function $\cP$, see Eq.~\eqref{eq_OrienPotential}. The parameter $\beta$ is a weighting parameter for the curvature term, see  Eq.~\eqref{eq_MetricInterpretation2}. In principle, $\beta$ should be tuned for each individual image. 
 
Finally, when applying the curvature-penalized metrics for the proposed model, we set the discretization resolution of the orientation dimension to be $60$. 

\begin{figure*}[t]
\centering
\includegraphics[width=17.5cm]{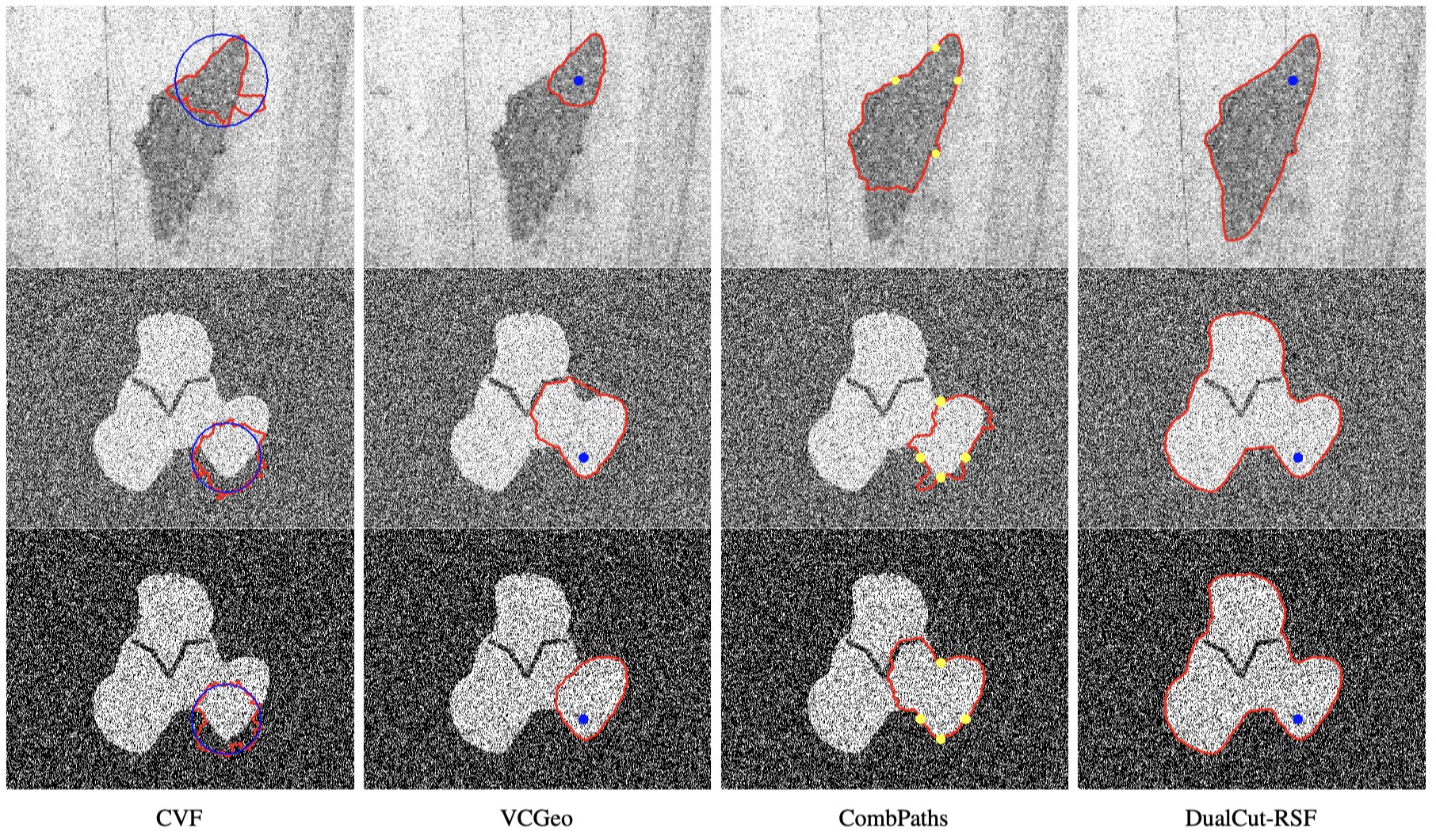}	
\caption{Qualitative comparison results with the CVF model, the VCGeo model and the CombPaths model, evaluated on images interrupted by different noises. The red lines indicate the final segmentation contours. The blue sold lines in column $1$ are the initial curves for the CVF model. The blue dots in columns $2$ and $4$ represent the landmark point $\fz$ for the VCGeo model and the proposed DualCut-RSF model, and the yellow dots in column $3$ are the input points for the CombPaths model}
\label{fig_NoisedImages}
\end{figure*}

\noindent\emph{Initialization for the considered models}.
The input of the proposed dual-cut model can be a  landmark point $\fz$ or a scribble $\cS$ inside the target region. In the following experiments, we exploit the point-based user input for the proposed model, unless otherwise noted. We make use of the point $\fz$ to initialize the VCGeo model for all the experiments. In addition, the initial contour for the VFC model is set as a circle centred at $\fz$ with a given radius. Finally, for the CombPaths model, we extract four control points clockwisely distributed along the ground truth boundary of interest. These points are sampled by taking $\fz$ as the origin of the image domain as before. Then the first control point is the intersection point between the positive $x$-axis and the ground truth boundary. Likewise to the first one,  the second to the fourth points are detected respectively using the negative $y$-axis, the negative $x$-axis and the positive $y$-axis. 
 
\subsection{Comparison Results}
One of the crucial contributions of the proposed dual-cut segmentation model lies at the introduction of the implicit region-based homogeneity information to guide the computation of geodesic paths. In contrast, the geodesic paths in both of the CombPaths and VCGeo models are dependent only on the edge-based features, which may yield bias toward the boundary segments of strong visibility, regardless of their Euclidean length. In Fig.~\ref{fig_syn}, we illustrate the effect from the region-based homogeneity terms in paths-based segmentation applications. In this test, the synthetic image used consists of two disjoint regions over the background. The region of interest lying at the bottom half of the image domain has highest gray levels. In Figs.~\ref{fig_syn}b and~\ref{fig_syn}c, the segmentation contours, as indicated by red lines, are generated from the CombPaths and VCGeo models, respectively. One can point out that each  of these segmentation contours combines a boundary segment not belonging to the target. In Figs.~\ref{fig_syn}d and~\ref{fig_syn}e, the contours are generated using the proposed DualCut-Asy and DualCut-RSF models. One can see that integrating the region-based homogeneity features with the image gradients coherence prior and the curvature regularization indeed can accurately capture the desired target region, as depicted in Figs.~\ref{fig_syn}d and~\ref{fig_syn}e.

\begin{table*}[t]
\centering
\caption{Quantitative comparisons between the VCF model, the VCGeo model,  the CombPaths model and the proposed DualCut-RSF model in terms of the statistics of Jaccard scores over $20$ runs per image. Images $1$ to $3$ are respectively shown from rows $1$ to $3$ of Fig.~\ref{fig_NoisedImages}}
\label{table_ResultNoise}
\setlength{\tabcolsep}{5pt}
\renewcommand{\arraystretch}{1.5}
\begin{tabular}[t]{c c c c c c c c c c c c c c c c c c}
\shline
\multicolumn{1}{c}{\multirow{1}{*}{Images}}&\multicolumn{4}{l}{CVF}&\multicolumn{4}{l}{VCGeo}&\multicolumn{4}{l}{CombPaths}&\multicolumn{4}{l}{DualCut-RSF} \\
\cmidrule(lr){2-5} \cmidrule(lr){6-9}\cmidrule(lr){10-13}\cmidrule(lr){14-17}
\multicolumn{1}{c}{}&\multicolumn{1}{c}{Mean}&\multicolumn{1}{c}{Max}&\multicolumn{1}{c}{Min}&\multicolumn{1}{l}{Std}&\multicolumn{1}{c}{Mean}&\multicolumn{1}{c}{Max}&\multicolumn{1}{c}{Min}&\multicolumn{1}{l}{Std}&\multicolumn{1}{c}{Mean}&\multicolumn{1}{c}{Max}&\multicolumn{1}{c}{Min}&\multicolumn{1}{l}{Std}&\multicolumn{1}{c}{Mean}&\multicolumn{1}{c}{Max}&\multicolumn{1}{c}{Min}&\multicolumn{1}{l}{Std}\\
\hline
\multicolumn{1}{c}{Image 1}&\multicolumn{1}{c}{$0.13$}&\multicolumn{1}{c}{$0.22$}&\multicolumn{1}{c}{$0.09$}&\multicolumn{1}{l}{$0.04$}
&\multicolumn{1}{c}{$0.59$}&\multicolumn{1}{c}{$0.83$}&\multicolumn{1}{c}{$0.10$}&\multicolumn{1}{l}{$0.29$}
&\multicolumn{1}{c}{$0.80$}&\multicolumn{1}{c}{$0.96$}&\multicolumn{1}{c}{$0.66$}&\multicolumn{1}{l}{$0.09$}
&\multicolumn{1}{c}{$0.96$}&\multicolumn{1}{c}{$0.96$}&\multicolumn{1}{c}{$0.93$}&\multicolumn{1}{l}{$\approx0$}\\
\multicolumn{1}{c}{Image 2}&\multicolumn{1}{c}{$0.17$}&\multicolumn{1}{c}{$0.25$}&\multicolumn{1}{c}{$0.10$}&\multicolumn{1}{l}{$0.05$}
&\multicolumn{1}{c}{$0.48$}&\multicolumn{1}{c}{$0.82$}&\multicolumn{1}{c}{$0.10$}&\multicolumn{1}{l}{$0.25$}
&\multicolumn{1}{c}{$0.55$}&\multicolumn{1}{c}{$0.97$}&\multicolumn{1}{c}{$0.01$}&\multicolumn{1}{l}{$0.42$}
&\multicolumn{1}{c}{$0.98$}&\multicolumn{1}{c}{$0.98$}&\multicolumn{1}{c}{$0.96$}&\multicolumn{1}{l}{$\approx 0$}\\

\multicolumn{1}{c}{Image 3}&\multicolumn{1}{c}{$0.16$}&\multicolumn{1}{c}{$0.23$}&\multicolumn{1}{c}{$0.10$}&\multicolumn{1}{l}{$0.04$}
&\multicolumn{1}{c}{$0.54$}&\multicolumn{1}{c}{$0.98$}&\multicolumn{1}{c}{$0.23$}&\multicolumn{1}{l}{$0.30$}
&\multicolumn{1}{c}{$0.83$}&\multicolumn{1}{c}{$0.98$}&\multicolumn{1}{c}{$0.32$}&\multicolumn{1}{l}{$0.24$}
&\multicolumn{1}{c}{$0.98$}&\multicolumn{1}{c}{$0.99$}&\multicolumn{1}{c}{$0.96$}&\multicolumn{1}{l}{$\approx0$}\\
\shline
\end{tabular}
\end{table*}

\begin{figure*}[t]
\centering
\includegraphics[width=17.5cm]{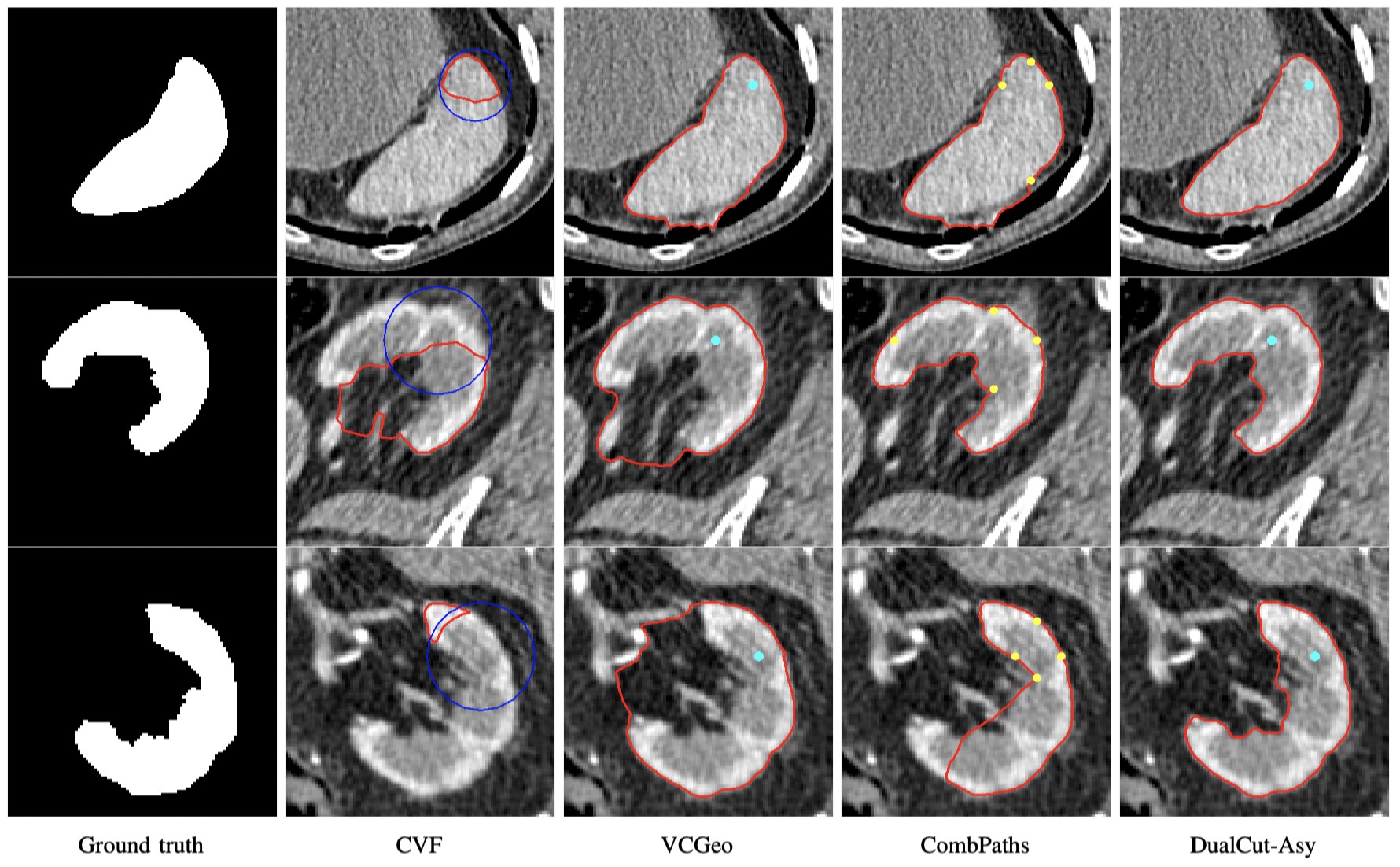}	
\caption{Qualitative comparison results on CT images with the CVF model, the VCGeo model and the CombPaths model. In column 1, we illustrate the ground truth regions. The red lines in columns $2$ to $4$ denote the segmentation contours derived from each model}
\label{fig_CT}	
\end{figure*}

\begin{figure}[t]
\centering
\includegraphics[width=8cm]{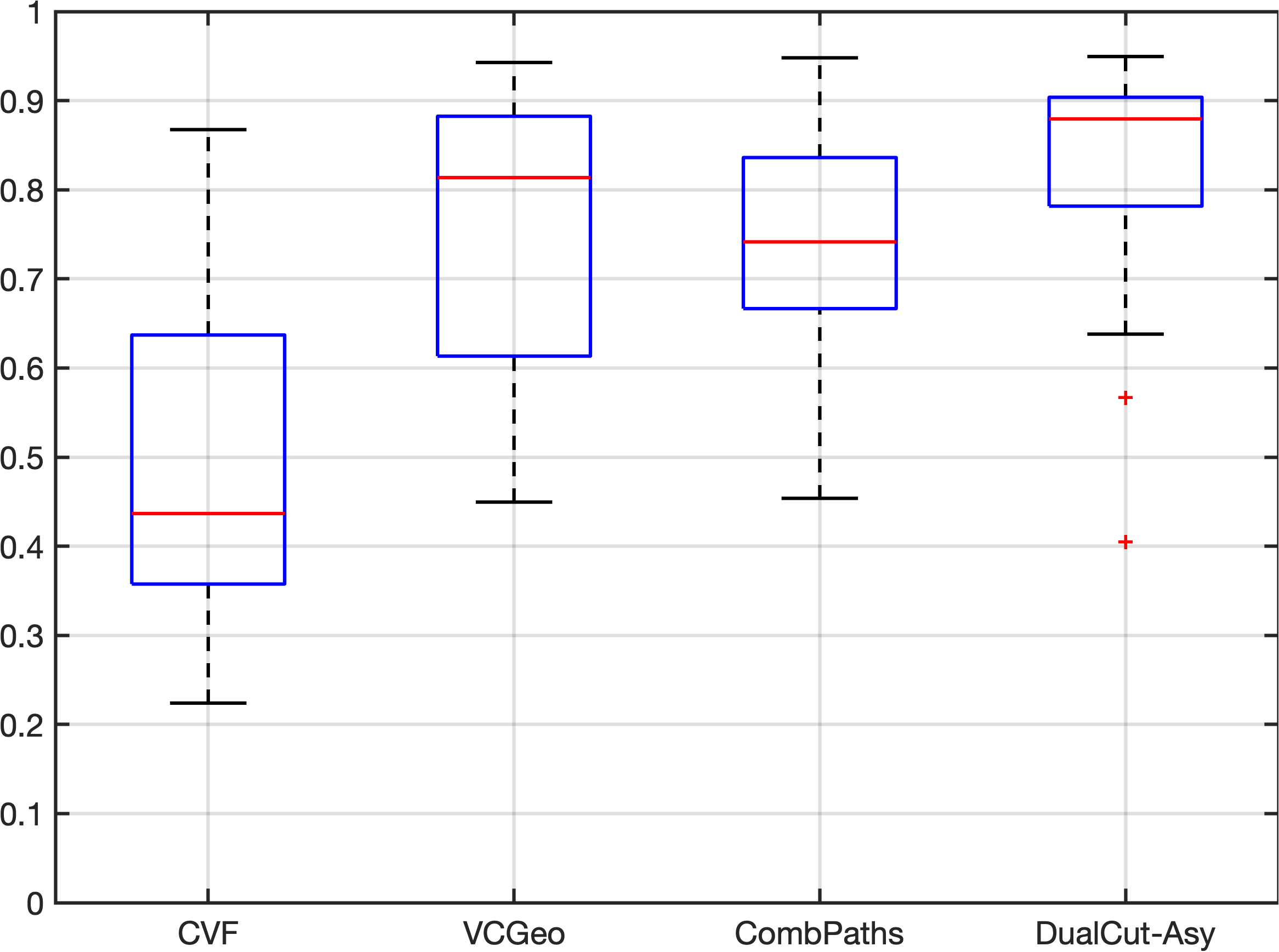}
\caption{Box plots of the average Jaccard scores of $20$ runs per image over a set of $86$ CT images.  Note that each test image is artificially interrupted by \emph{additive Gaussian noises}}
\label{fig_boxplot_CT}
\end{figure}

In Fig.~\ref{fig_examples_natural}, we demonstrate  the qualitative results for the CVF model, the VCGeo model, the CombPaths model and the DualCut-Asy model on real images sampled from the Grabcut dataset~\cite{rother2004grabcut} and the Weizmann dataset~\cite{alpert2012image}. In this experiment, the segmentation contours of each evaluated model are denoted by red lines. The white lines in row $2$ represent the initial curves for CVF model. The blue (resp. yellow) dots in rows $2$ and $4$ (resp. row $3$) are the input points for the VCGeo model and the proposed DualCut-Asy model (resp. the CombPaths model), respectively. From Fig.~\ref{fig_examples_natural}, we can see that most of the segmentation contours derived  from the CVF, VCGeo and CombPaths models suffer from the shortcut problems, as depicted in rows $1$ to $3$. In other words, most of these segmentation contours  pass through the interior of the target regions. The DualCut-Asy model in conjunction with the implicit region-based homogeneity features is able to reduce the risk for the segmentation curves  being trapped into unexpected local minima, as shown in row $4$. Note that for the proposed DualCut-Asy model, we exploit the piecewise constants-based homogeneity term to derive $\psi_\fz$ for the test images in the first two rows and the  Bhattacharyya coefficients for the images in rows $3$ to $5$. 

In Fig.~\ref{fig_examples_obstacle}, we demonstrate the advantages of using the performance of the proposed DualCut-Asy model on an image with complicated foreground, when exploiting scribbles as input. For this purpose, we first sample $20$ seed points from the interior of the ground truth region, as indicated by the cyan dots in Fig.~\ref{fig_examples_obstacle}a. From each seed point, we perform the front propagation associated to a potential with constant value $1$ inside the eroded ground truth region and $\infty$, otherwise.  A farthest point is a point which has the highest distance value among all the boundary points of the eroded ground truth region. From this farthest point, one can track a shortest path linking to the corresponding seed.  The scribbles of the second type are provided manually, as depicted by the white lines in Fig.~\ref{fig_examples_obstacle}c. With these setting, we perform $20$ runs for all the evaluated models such that each seed point is taken as the landmark point to set up the respective models. Note that in each test, we use the scribble of both types for the VCGeo model and the DualCut-Asy model.
The statistics for the corresponding segmentation results are illustrated in Fig.~\ref{fig_examples_obstacle}c. From the box plots, we can see that the DualCut-Asy model achieves the best performance in the sense of Jaccard scores, due to the benefits from both of the region-based homogeneity enhancement and image gradients coherence property.  Even through with the same scribbles during geodesic paths estimation, the segmentation results from the VCGeo model show lower accuracy than the proposed one. The segmentation results from the CombPaths model seem to be insensitive to the input. However, we still observe lower Jaccard scores comparing to the proposed DualCut-Asy model. The Jaccard scores for the CVF model illustrate that the segmentations from this model in most tests might be stuck in unexpected local minima. In this experiment, we use the  Bhattacharyya coefficients for computing $\psi_\fz$. The average computation time of the DualCut-Asy model is around $1$ second involving the computation of the velocity $\xi$, the construction of the  metrics and the computation of geodesic distances, where the test image involves $480\times 640$ grid points.

The curvature regularization may lead to smooth geodesic paths. This property is very effective in scenarios of segmenting images with strong noise levels. We show the comparison results on three images interrupted by different noises, as depicted in Fig.~\ref{fig_NoisedImages}. Moreover, in Table~\ref{table_ResultNoise}, we illustrate the statistics of the Jaccard scores with respect to different image segmentation approaches over $20$ runs per image, using a set of sampled points distributed inside the regions of interest. In this experiment, the piecewise constants-based homogeneity term is used for building $\psi_\fz$. The results illustrated in this table indeed show that the proposed DualCut-RSF model are able to capture favorable segmentations in the presence of strong noises. For the images in the rows $2$ and $3$, the computation time of the proposed DualCut-RSF model is around $8$ seconds  where the resolution of the grid  $\bM_h$ is $450\times 600\times 60$. 

In Fig.~\ref{fig_CT}, we perform the qualitative comparison on three CT images, for which the image gradients coherence property is satisfied along most parts of the objective boundaries. In this experiment, one can observe that the CVF model fails to segment the whole object regions. Moreover,  the CombPaths model is capable of roughly capturing the boundaries of interest as depicted in rows $1$ to $3$. In rows $2$ and $3$, we observe leakage problems for the VCGeo model, where a portion of these segmentation contour pass through the background region. In contrast, the proposed DualCut-Asy model is able to find favorable segmentations in all CT images, due to the benefits from the regional homogeneity features derived using piecewise constants-based term in this experiment, and from the use of the image gradients coherence prior. Finally, we present the quantitative comparison results over $86$ CT images~\cite{spencer2019parameter} in terms of Jaccard scores. Each test CT image is artificially interrupted by Gaussian white noise with mean $0$ and normalized variance $0.05$.  We sample $20$ points for each image from the corresponding eroded ground truth region.  Each sampled point is taken as the landmark point $\fz$ to initialize the evaluated models. The box plots of the mean Jaccard scores over $20$ runs are illustrated in Fig.~\ref{fig_boxplot_CT}. In each test, two and four points lying at the ground truth boundary are exploit to set up the VCGeo model and the CombPaths model.
One can see that the proposed DualCut-Asy model indeed outperforms the compared models, even through the VCGeo model and the CombPaths model make use of more reliable user intervention. 

\noindent
\emph{Discussion on future work}.
In summary, the proposed geodesic paths-based model requires a point $\fz$ to set up the initialization. In this work, we assume that the point $\fz$ is provided by user, yielding an efficient and accurate interactive segmentation algorithm. In the future, the research work can be devoted to the integration with learning-based approaches. A possible way is to exploit the saliency objection detection models~\cite{han2018advanced,zhang2019synthesizing,wei2012geodesic,wei2020f3net} to predict the position of $\fz$ and also to take the saliency maps as auxiliary image features. In addition, we plan to take into account the shape priors of convexity and star-convexity  for the computation of geodesic paths.

\section{Conclusion}
\label{sec_Conclusion}
In this paper, an efficient geodesic paths-based model is introduced for  interactive image segmentation under the Eikonal PDE framework. We show the possibility of implicitly incorporating a great variety of  region-based homogeneity information into the construction of local geodesic metrics. The implicit representation of the region-based features is carried out by a scalar-valued function, which is used the weight for a Finsler geodesic metric. As a consequence, either the image gradients coherence prior or the curvature regularization  can be taken into account for tracking geodesic paths. In addition, we also introduce a new dual-cut geodesic computation scheme for image segmentation using a landmark point inside the target region.  Both of the qualitative and quantitive comparison results prove the advantages of the proposed model in interactive image segmentation. 

\section*{Acknowledgment}
The authors thank all the anonymous reviewers for their invaluable  suggestions to improve this manuscript. This work is in part supported by the National Natural Science Foundation of China (No. 61902224), the Shandong Provincial Natural Science Foundation (No. ZR2020LZL001), and the French government under management of Agence Nationale de la Recherche as part of the ``Investissements d'avenir'' program, reference ANR-19-P3IA-0001 (PRAIRIE 3IA Institute). This research is also partially supported by the Distinguished Taishan Scholars in Climbing Plan (No. tspd20181211) and Young Taishan Scholars (Nos. tsqn201909137, tsqn201909140).

\appendix
\label{sec:appendix}

\subsection{The VCGeo Model}
\label{appendix_VCGeo}
We consider a variant of the original circular geodesic model~\cite{appleton2005globally}, named VCGeo model,  as a baseline algorithm.  This model is implemented by a disjoint paths concatenation scheme, where the final segmentation is a region involves a given point $\fz$. We apply the anisotropic Riemannian metric $\cF^{\rm R}_\fz:\Omega\backslash\{\fz\}\times\bR^2\to\bR^+_0$ weighted by the balloon term for the VCGeo model
\begin{equation}
\label{eq_MetricVCGeo}
\cF^{\rm R}_\fz(\fx,\fu)=\|\fx-\fz\|^{-1}\sqrt{\langle\fu,\cM(\fx)\fu\rangle},
\end{equation}
where the tensor field $\cM$ is defined in~\eqref{eq_EdgeM}.

As in the proposed dual-cut model, the point $\fz$ is instantiated as the origin of the image domain $\Omega$. Assume that $\cC_{\rm GT}$ with $\cC_{\rm GT}(0)\in\ell_\fz^-$ is a closed curve defined over the range $[0,1]$ that counter-clockwisely parameterizes the boundary of the ground truth region. In this case, one can sample the \emph{first} and \emph{last} intersection points  between $\ell^+_\fz$ and $\cC_{\rm GT}$, respectively denoted by $\cC_{\rm GT}(u_1)=\fa$ and $\cC_{\rm GT}(u_2)=\fb$ with $u_1\leq u_2$. In case $\fa\neq\fb$, the target closed curve $\cC$ is taken as the concatenation of two geodesic paths $\cG^-_{\fb,\fa}$ and $\cG^+_{\fa,\fb}$ using Eq.~\eqref{eq_Concatenation}. The first geodesic path  $\cG^-_{\fb,\fa}$ linking from $\fb$ to $\fa$ is tracked using the HFM method, where the offsets $\fe_j$ are removed from the stencil $\Lambda(\fx)$ if they satisfy the Points (i) and (ii) with respect to $\ell_\fz^+$,  see Section~\ref{subsec_NI}. Likewise the computation of $\cG^-_{\fb,\fa}$,  the second geodesic path $\cG^+_{\fa,\fb}$ is generated by the HFM method using the constrained stencils with respect to  $\ell^-_\fz$. Finally, in case $\fa=\fb$, we set the final curve as $\cC=\cG_{\fb,\fa}^-$.

\ifCLASSOPTIONcaptionsoff
 \newpage
\fi

\bibliographystyle{IEEEbib}
\bibliography{minimalPaths}

\begin{thebibliography}{10}

\bibitem{boykov2006grapy}
Y.~Boykov and G.~Funka-Lea,
\newblock ``{Graph cuts and efficient N-D image segmentation},''
\newblock {\em Int. J. Comput. Vis.}, vol. 70, pp. 109–--131, 2006.

\bibitem{grady2006random}
L.~Grady,
\newblock ``Random walks for image segmentation,''
\newblock {\em IEEE Trans. Pattern Anal. Mach. Intell.}, vol. 28, no. 11, pp.
  1768--1783, 2006.

\bibitem{couprie2011power}
C.~{Couprie}, L.~{Grady}, L.~{Najman}, and H.~{Talbot},
\newblock ``{Power watershed: A unifying graph-based optimization framework},''
\newblock {\em IEEE Trans. Pattern Anal. Mach. Intell.}, vol. 33, no. 7, pp.
  1384--1399, 2011.

\bibitem{li2004lzay}
Y.~Li, J.~Sun, C.-K. Tang, and H.-Y. Shum,
\newblock ``Lazy snapping,''
\newblock {\em ACM Trans. Graph.}, vol. 23, no. 3, pp. 303–308, 2004.

\bibitem{arbelaez2004energy}
P.~A. Arbel{\'a}ez and L.~D Cohen,
\newblock ``Energy partitions and image segmentation,''
\newblock {\em J. Math. Imaging Vis.}, vol. 20, no. 1, pp. 43--57, 2004.

\bibitem{bai2009geodesic}
X.~Bai and G.~Sapiro,
\newblock ``Geodesic matting: A framework for fast interactive image and video
  segmentation and matting,''
\newblock {\em Int. J. Comput. Vis.}, vol. 82, no. 2, pp. 113--132, 2009.

\bibitem{chen2018fast}
D.~Chen and L.~D. Cohen,
\newblock ``Fast asymmetric fronts propagation for image segmentation,''
\newblock {\em J. Math. Imaging Vis.}, vol. 60, no. 6, pp. 766--783, 2018.

\bibitem{gao2012interactive}
Y.~Gao, R.~Kikinis, S.~Bouix, M.~Shenton, and A.~Tannenbaum,
\newblock ``{A 3D interactive multi-object segmentation tool using local robust
  statistics driven active contours},''
\newblock {\em Med. Image Anal.}, vol. 16, no. 6, pp. 1216--1227, 2012.

\bibitem{spencer2019parameter}
J.~{Spencer}, K.~{Chen}, and J.~{Duan},
\newblock ``Parameter-free selective segmentation with convex variational
  methods,''
\newblock {\em IEEE Trans. Image Process.}, vol. 28, no. 5, pp. 2163--2172,
  2019.

\bibitem{nguyen2012robust}
T.~N.~A. Nguyen, J.~Cai, J.~Zhang, and J.~Zheng,
\newblock ``Robust interactive image segmentation using convex active
  contours,''
\newblock {\em IEEE Trans. Image Process.}, vol. 21, no. 8, pp. 3734--3743,
  2012.

\bibitem{kass1988snakes}
M.~Kass, A.~Witkin, and D.~Terzopoulos,
\newblock ``{Snakes: Active contour models},''
\newblock {\em Int. J. Comput. Vis.}, vol. 1, no. 4, pp. 321--331, 1988.

\bibitem{xu1998snakes}
C.~Xu and J.~L. Prince,
\newblock ``Snakes, shapes, and gradient vector flow,''
\newblock {\em IEEE Trans. Image Process.}, vol. 7, no. 3, pp. 359--369, 1998.

\bibitem{cohen1991active}
L.~D. Cohen,
\newblock ``On active contour models and balloons,''
\newblock {\em CVGIP: Image Understand.}, vol. 53, no. 2, pp. 211--218, 1991.

\bibitem{cohen1993finite}
L.~D. Cohen and I.~Cohen,
\newblock ``{Finite-element methods for active contour models and balloons for
  2-D and 3-D images},''
\newblock {\em IEEE Trans. Pattern Anal. Mach. Intell.}, vol. 15, no. 11, pp.
  1131--1147, 1993.

\bibitem{caselles1997geodesic}
V.~Caselles, R.~Kimmel, and G.~Sapiro,
\newblock ``Geodesic active contours,''
\newblock {\em Int. J. Comput. Vis.}, vol. 22, no. 1, pp. 61--79, 1997.

\bibitem{malladi1995shape}
R.~Malladi, J.~Sethian, and B.~C. Vemuri,
\newblock ``{Shape modeling with front propagation: A level set approach},''
\newblock {\em IEEE Trans. Pattern Anal. Mach. Intell.}, vol. 17, no. 2, pp.
  158--175, 1995.

\bibitem{li2008minimization}
C.~Li, C.~Kao, J.~C. Gore, and Z.~Ding,
\newblock ``Minimization of region-scalable fitting energy for image
  segmentation,''
\newblock {\em IEEE Trans. Image Process.}, vol. 17, no. 10, pp. 1940--1949,
  2008.

\bibitem{brox2009local}
T.~Brox and D.~Cremers,
\newblock ``{On local region models and a statistical interpretation of the
  piecewise smooth Mumford-Shah functional},''
\newblock {\em Int. J. Comput. Vis.}, vol. 84, no. 2, pp. 184--193, 2009.

\bibitem{miranda2012riverbed}
P.~A.~V. Miranda, A.~X. Falcao, and T.~V. Spina,
\newblock ``{Riverbed: A novel user-steered image segmentation method based on
  optimum boundary tracking},''
\newblock {\em IEEE Trans. Image Process.}, vol. 21, no. 6, pp. 3042--3052,
  2012.

\bibitem{cohen1997global}
L.~D. Cohen and R.~Kimmel,
\newblock ``{Global minimum for active contour models: A minimal path
  approach},''
\newblock {\em Int. J. Comput. Vis.}, vol. 24, no. 1, pp. 57--78, 1997.

\bibitem{cremers2002diffusion}
D.~Cremers, F.~Tischh{\"a}user, J.~Weickert, and C.~Schn{\"o}rr,
\newblock ``{Diffusion snakes: Introducing statistical shape knowledge into the
  Mumford-Shah functional},''
\newblock {\em Int. J. Comput. Vis.}, vol. 50, no. 3, pp. 295--313, 2002.

\bibitem{bresson2006variational}
X.~Bresson, P.~Vandergheynst, and J.-P. Thiran,
\newblock ``{A variational model for object segmentation using boundary
  information and shape prior driven by the Mumford-Shah functional},''
\newblock {\em Int. J. Comput. Vis.}, vol. 68, no. 2, pp. 145--162, 2006.

\bibitem{leventon2000statistical}
M.~E. {Leventon}, W.~E.~L. {Grimson}, and O.~{Faugeras},
\newblock ``Statistical shape influence in geodesic active contours,''
\newblock in {\em Proc. CVPR}, 2000, vol.~1, pp. 316--323.

\bibitem{chan2005level}
T.~Chan and W.~Zhu,
\newblock ``Level set based shape prior segmentation,''
\newblock in {\em Proc. CVPR}. IEEE, 2005, vol.~2, pp. 1164--1170.

\bibitem{klodt2011convex}
M.~Klodt and D.~Cremers,
\newblock ``A convex framework for image segmentation with moment
  constraints,''
\newblock in {\em Proc. ICCV}. IEEE, 2011, pp. 2236--2243.

\bibitem{yan2020convexity}
S.~Yan, X.-C. Tai, J.~Liu, and H.-Y. Huang,
\newblock ``Convexity shape prior for level set-based image segmentation
  method,''
\newblock {\em IEEE Trans. Image Process.}, vol. 29, pp. 7141--7152, 2020.

\bibitem{luo2019convex}
S.~Luo, X.-C. Tai, L.~Huo, Y.~Wang, and R.~Glowinski,
\newblock ``Convex shape prior for multi-object segmentation using a single
  level set function,''
\newblock in {\em Proc. CVPR}, 2019, pp. 613--621.

\bibitem{gorelick2016convexity}
L.~Gorelick, O.~Veksler, Y.~Boykov, and C.~Nieuwenhuis,
\newblock ``Convexity shape prior for binary segmentation,''
\newblock {\em IEEE Trans. Pattern Anal. Mach. Intell.}, vol. 39, no. 2, pp.
  258--271, 2016.

\bibitem{royer2016convexity}
L.~A. Royer, D.~L. Richmond, C.~Rother, B.~Andres, and D.~Kainmueller,
\newblock ``Convexity shape constraints for image segmentation,''
\newblock in {\em Proc. CVPR}, 2016, pp. 402--410.

\bibitem{veksler2008star}
O.~Veksler,
\newblock ``Star shape prior for graph-cut image segmentation,''
\newblock in {\em Proc. ECCV}. Springer, 2008, pp. 454--467.

\bibitem{vicente2008graph}
S.~Vicente, V.~Kolmogorov, and C.~Rother,
\newblock ``Graph cut based image segmentation with connectivity priors,''
\newblock in {\em Proc. CVPR}, 2008.

\bibitem{yuan2012efficient}
J.~Yuan, W.~Qiu, E.~Ukwatta, M.~Rajchl, Y.~Sun, and A.~Fenster,
\newblock ``{An efficient convex optimization approach to 3D prostate MRI
  segmentation with generic star shape prior},''
\newblock {\em Prostate MR Image Segmentation Challenge, MICCAI}, vol. 7512,
  pp. 82--89, 2012.

\bibitem{ma2020learning}
J.~Ma, J.~He, and X.~Yang,
\newblock ``Learning geodesic active contours for embedding object global
  information in segmentation cnns,''
\newblock {\em IEEE Trans. on Medical Imaging}, 2020.

\bibitem{wang2018deepigeos}
G.~Wang, M.~A. Zuluaga, W.~Li, R.~Pratt, P.~A Patel, M.~Aertsen, T.~Doel, A.~L
  David, J.~Deprest, S.~Ourselin, et~al.,
\newblock ``Deepigeos: a deep interactive geodesic framework for medical image
  segmentation,''
\newblock {\em IEEE Trans. Pattern Anal. Mach. Intell.}, vol. 41, no. 7, pp.
  1559--1572, 2018.

\bibitem{zhang2020exploring}
D.~Zhang, G.~Huang, Q.~Zhang, J.~Han, J.~Han, Y.~Wang, and Y.~Yu,
\newblock ``{Exploring task structure for brain tumor segmentation from
  multi-modality MR images},''
\newblock {\em IEEE Trans. Image Process.}, vol. 29, pp. 9032--9043, 2020.

\bibitem{zhang2021automatic}
D.~Zhang, J.~Zhang, Q.~Zhang, J.~Han, S.~Zhang, and J.~Han,
\newblock ``Automatic pancreas segmentation based on lightweight dcnn modules
  and spatial prior propagation,''
\newblock {\em Pattern Recognition}, vol. 114, pp. 107762, 2021.

\bibitem{zhang2021cross}
D.~Zhang, G.~Huang, Q.~Zhang, J.~Han, J.~Han, and Y.~Yu,
\newblock ``Cross-modality deep feature learning for brain tumor
  segmentation,''
\newblock {\em Pattern Recognition}, vol. 110, pp. 107562, 2021.

\bibitem{appleton2005globally}
B.~Appleton and H.~Talbot,
\newblock ``Globally optimal geodesic active contours,''
\newblock {\em J. Math. Imaging Vis.}, vol. 23, no. 1, pp. 67--86, 2005.

\bibitem{yezzi1997geometric}
A.~Yezzi, S.~Kichenassamy, A.~Kumar, P.~Olver, and A.~Tannenbaum,
\newblock ``A geometric snake model for segmentation of medical imagery,''
\newblock {\em IEEE Trans. Med. Imaging}, vol. 16, no. 2, pp. 199--209, 1997.

\bibitem{kimmel2003regularized}
R.~Kimmel and A.~M. Bruckstein,
\newblock ``Regularized laplacian zero crossings as optimal edge integrators,''
\newblock {\em Int. J. Comput. Vis.}, vol. 53, no. 3, pp. 225--243, 2003.

\bibitem{osher1988fronts}
S.~Osher and J.~A. Sethian,
\newblock ``{Fronts propagating with curvature-dependent speed: algorithms
  based on Hamilton-Jacobi formulations},''
\newblock {\em J. Comput. Phys.}, vol. 79, no. 1, pp. 12--49, 1988.

\bibitem{peyre2010geodesic}
G.~Peyr{\'e}, M.~P{\'e}chaud, R.~Keriven, and L.~D. Cohen,
\newblock ``Geodesic methods in computer vision and graphics,''
\newblock {\em Foundations and Trends{\textregistered} in Computer Graphics and
  Vision}, vol. 5, no. 3--4, pp. 197--397, 2010.

\bibitem{benmansour2009fast}
F.~Benmansour and L.~D. Cohen,
\newblock ``{Fast object segmentation by growing minimal paths from a single
  point on 2D or 3D images},''
\newblock {\em J. Math. Imaging Vis.}, vol. 33, no. 2, pp. 209--221, 2009.

\bibitem{mille2009geodesically}
J.~Mille and Laurent~D Cohen,
\newblock ``Geodesically linked active contours: evolution strategy based on
  minimal paths,''
\newblock in {\em Proc. SSVM}. Springer, 2009, pp. 163--174.

\bibitem{mille2015combination}
J.~Mille, S.~Bougleux, and L.~D. Cohen,
\newblock ``Combination of piecewise-geodesic paths for interactive
  segmentation,''
\newblock {\em Int. J. Comput. Vis.}, vol. 112, no. 1, pp. 1--22, 2015.

\bibitem{appia2011active}
V.~Appia and A.~Yezzi,
\newblock ``{Active geodesics: Region-based active contour segmentation with a
  global edge-based constraint},''
\newblock in {\em Proc. ICCV}. IEEE, 2011, pp. 1975--1980.

\bibitem{chen2016finsler}
D.~Chen, J.-M. Mirebeau, and L.~D. Cohen,
\newblock ``Finsler geodesics evolution model for region based active
  contours,''
\newblock in {\em Proc. BMVC}, 2016.

\bibitem{chen2019active}
D.~Chen and L.~D. Cohen,
\newblock ``{From active contours to minimal geodesic paths: New solutions to
  active contours problems by Eikonal equations},''
\newblock in {\em Handbook of Numerical Analysis}, vol.~20, pp. 233--271.
  Elsevier, 2019.

\bibitem{melonakos2008finsler}
J.~Melonakos, E.~Pichon, S.~Angenent, and A.~Tannenbaum,
\newblock ``Finsler active contours,''
\newblock {\em IEEE Trans. Pattern Anal. Mach. Intell.}, vol. 30, no. 3, pp.
  412--423, 2008.

\bibitem{randers1941asymmetrical}
G.~Randers,
\newblock ``On an asymmetrical metric in the four-space of general
  relativity,''
\newblock {\em Phys. Rev.}, vol. 59, no. 2, pp. 195, 1941.

\bibitem{chen2017global}
D.~Chen, J.-M. Mirebeau, and L.~D. Cohen,
\newblock ``{Global minimum for a Finsler elastica minimal path approach},''
\newblock {\em Int. J. Comput. Vis.}, vol. 122, no. 3, pp. 458--483, 2017.

\bibitem{chen2018asymmetric}
D.~Chen, J.~Spencer, J.~M. Mirebeau, K.~Chen, and L.~D. Cohen,
\newblock ``{Asymmetric geodesic distance propagation for active contours},''
\newblock in {\em Proc. BMVC}, 2018.

\bibitem{duits2018optimal}
R.~Duits, S.~PL Meesters, J.-M. Mirebeau, and J.~M Portegies,
\newblock ``{Optimal paths for variants of the 2D and 3D Reeds--Shepp car with
  applications in image analysis},''
\newblock {\em J. Math. Imag. Vis.}, vol. 60, no. 6, pp. 816--848, 2018.

\bibitem{mirebeau2018fast}
J.-M. Mirebeau,
\newblock ``Fast-marching methods for curvature penalized shortest paths,''
\newblock {\em J. Math. Imag. Vis.}, vol. 60, no. 6, pp. 784--815, 2018.

\bibitem{mirebeau2014efficient}
J.-M. Mirebeau,
\newblock ``{Efficient fast marching with Finsler metrics},''
\newblock {\em Numer. Math.}, vol. 126, no. 3, pp. 515--557, 2014.

\bibitem{mirebeau2019riemannian}
J.-M. Mirebeau,
\newblock ``{Riemannian fast-marching on Cartesian grids, using Voronoi's first
  reduction of quadratic forms},''
\newblock {\em SIAM J. Numer. Anal.}, vol. 57, no. 6, pp. 2608--2655, 2019.

\bibitem{chan2001active}
T.~F. Chan and L.~A. Vese,
\newblock ``Active contours without edges,''
\newblock {\em IEEE Trans. Image Process.}, vol. 10, no. 2, pp. 266--277, 2001.

\bibitem{cremers2007review}
D.~Cremers, M.~Rousson, and R.~Deriche,
\newblock ``A review of statistical approaches to level set segmentation:
  integrating color, texture, motion and shape,''
\newblock {\em Int. J. Comput. Vis.}, vol. 72, no. 2, pp. 195--215, 2007.

\bibitem{zhu1996region}
S.~Zhu and A.~Yuille,
\newblock ``{Region competition: Unifying snakes, region growing, and Bayes/MDL
  for multiband image segmentation},''
\newblock {\em IEEE Trans. Pattern Anal. Mach. Intell.}, vol. 18, no. 9, pp.
  884--900, 1996.

\bibitem{dubrovina2015multi}
A.~Dubrovina-Karni, G.~Rosman, and R.~Kimmel,
\newblock ``Multi-region active contours with a single level set function,''
\newblock {\em IEEE Trans. Pattern Anal. Mach. Intell.}, vol. 37, no. 8, pp.
  1585--1601, 2015.

\bibitem{jung2012nonlocal}
M.~Jung, G.~Peyr{\'e}, and L.~D. Cohen,
\newblock ``Nonlocal active contours,''
\newblock {\em SIAM J. Imaging Sci.}, vol. 5, no. 3, pp. 1022--1054, 2012.

\bibitem{sumengen2006graph}
B.~Sumengen and BS~Manjunath,
\newblock ``{Graph partitioning active contours (GPAC) for image
  segmentation},''
\newblock {\em IEEE Trans. Pattern Anal. Mach. Intell.}, vol. 28, no. 4, pp.
  509--521, 2006.

\bibitem{michailovich2007image}
O.~Michailovich, Y.~Rathi, and A.~Tannenbaum,
\newblock ``{Image segmentation using active contours driven by the
  Bhattacharyya gradient flow},''
\newblock {\em IEEE Trans. Image Process.}, vol. 16, no. 11, pp. 2787--2801,
  2007.

\bibitem{chan2000active}
T.~F. Chan, B.~Y. Sandberg, and L.~A. Vese,
\newblock ``{Active contours without edges for vector-valued images},''
\newblock {\em J. Vis. Commun. Image Represent.}, vol. 11, no. 2, pp. 130--141,
  2000.

\bibitem{sochen1998vision}
N.~Sochen, R.~Kimmel, and R.~Malladi,
\newblock ``A general framework for low level vision,''
\newblock {\em IEEE Trans. Image Process.}, vol. 7, no. 3, pp. 310--318, 1998.

\bibitem{tsai2001curve}
A.~Tsai, A.~Yezzi, and A.~S. Willsky,
\newblock ``{Curve evolution implementation of the Mumford-Shah functional for
  image segmentation, denoising, interpolation, and magnification},''
\newblock {\em IEEE Trans. Image Process.}, vol. 10, no. 8, pp. 1169--1186,
  2001.

\bibitem{chen2019eikonal}
D.~Chen, J.-M. Mirebeau, H.~Shu, and L.~D. Cohen,
\newblock ``Eikonal region-based active contours for image segmentation,''
\newblock {\em arXiv preprint arXiv:1912.10122}, 2019.

\bibitem{zach2009globally}
C.~Zach, L.~Shan, and M.~Niethammer,
\newblock ``{Globally optimal Finsler active contours},''
\newblock in {\em Proc. Joint Pattern Recognition Symposium}. Springer, 2009,
  pp. 552--561.

\bibitem{li2007active}
B.~Li and S.~T. Acton,
\newblock ``Active contour external force using vector field convolution for
  image segmentation,''
\newblock {\em IEEE Trans. Image Process.}, vol. 16, no. 8, pp. 2096--2106,
  2007.

\bibitem{rother2004grabcut}
C.~Rother, V.~Kolmogorov, and A.~Blake,
\newblock ``{Grabcut: Interactive foreground extraction using iterated graph
  cuts},''
\newblock {\em ACM Trans. Graph.}, vol. 23, no. 3, pp. 309--314, 2004.

\bibitem{alpert2012image}
S.~Alpert, M.~Galun, A.~Brandt, and R.~Basri,
\newblock ``Image segmentation by probabilistic bottom-up aggregation and cue
  integration,''
\newblock {\em IEEE Trans. Pattern Anal. Mach. Intell.}, vol. 34, no. 2, pp.
  315--327, 2012.

\bibitem{han2018advanced}
J.~Han, D.~Zhang, G.~Cheng, N.~Liu, and D.~Xu,
\newblock ``Advanced deep-learning techniques for salient and category-specific
  object detection: a survey,''
\newblock {\em IEEE Signal Process. Mag.}, vol. 35, no. 1, pp. 84--100, 2018.

\bibitem{zhang2019synthesizing}
D.~Zhang, J.~Han, Y.~Zhang, and D.~Xu,
\newblock ``Synthesizing supervision for learning deep saliency network without
  human annotation,''
\newblock {\em IEEE Trans. Pattern Anal. Mach. Intell.}, vol. 42, no. 7, pp.
  1755--1769, 2019.

\bibitem{wei2012geodesic}
Y.~Wei, F.~Wen, W.~Zhu, and J.~Sun,
\newblock ``Geodesic saliency using background priors,''
\newblock in {\em Proc. ECCV}. Springer, 2012, pp. 29--42.

\bibitem{wei2020f3net}
J.~Wei, S.~Wang, and Q.~Huang,
\newblock ``{F$^3$Net: fusion, feedback and focus for salient object
  detection},''
\newblock in {\em Proc. AAAI}, 2020, vol.~34, pp. 12321--12328.

\end{thebibliography}

\end{document}